\documentclass{article}

\usepackage{arxiv}

\usepackage[utf8]{inputenc} % allow utf-8 input
\usepackage[T1]{fontenc}    % use 8-bit T1 fonts
\usepackage{hyperref}       % hyperlinks
\usepackage{url}            % simple URL typesetting
\usepackage{booktabs}       % professional-quality tables
\usepackage{amsfonts}       % blackboard math symbols
\usepackage{nicefrac}       % compact symbols for 1/2, etc.
\usepackage{microtype}      % microtypography
\usepackage{lipsum}
\usepackage{subfigure}
\usepackage{graphicx}
\usepackage{amsmath}
\title{Aerodynamic Data Predictions Based on Multi-task Learning}

\author{
  Liwei~Hu\\
  School of Computer Science and Engineering\\
  University  of  Electronic  Science and Technology of China\\
  Chengdu, Sichuan, 611731, China\\
  \And
  Yu~Xiang\thanks{Corresponding Author, Professor}\\
  School of Computer Science and Engineering\\
  University  of  Electronic  Science and Technology of China\\
  Chengdu, Sichuan, 611731, China\\
  \texttt{jcxiang@uestc.edu.cn} \\
  \AND
  Jun~Zhang\\
  School of Computer Science and Engineering\\
  University  of  Electronic  Science and Technology of China\\
  Chengdu, Sichuan, 611731, China\\
  \AND
  Zifang~Shi\\
  School of Computer Science and Engineering\\
  University  of  Electronic  Science and Technology of China\\
  Chengdu, Sichuan, 611731, China\\
  \AND
  Wenzheng~Wang\\
  China Aerodynamics Research and Development Center\\
  Mianyang, Sichuna, 621000, China\\
}

\begin{document}
\maketitle

\begin{abstract}
The quality of datasets is one of the key factors that affect the accuracy of aerodynamic data models. For example, in the uniformly sampled Burgers' dataset, the insufficient high-speed data is overwhelmed by massive low-speed data. Predicting high-speed data is more difficult than predicting low-speed data, owing to that the number of high-speed data is limited, i.e. the quality of the Burgers' dataset is not satisfactory. To improve the quality of datasets, traditional methods usually employ the data resampling technology to produce enough data for the insufficient parts in the original datasets before modeling, which increases computational costs. Recently, the mixtures of experts have been used in natural language processing to deal with different parts of sentences, which provides a solution for eliminating the need for data resampling in aerodynamic data modeling. Motivated by this, we propose the multi-task learning (MTL), a datasets quality-adaptive learning scheme, which combines task allocation and aerodynamic characteristics learning together to disperse the pressure of the entire learning task. The task allocation divides a whole learning task into several independent subtasks, while the aerodynamic characteristics learning learns these subtasks simultaneously to achieve better precision. Two experiments with poor quality datasets are conducted to verify the data quality-adaptivity of the MTL to datasets. The results show than the MTL is more accurate than FCNs and GANs in poor quality datasets.
\end{abstract}

% keywords can be removed

\section{Introduction}
\begin{figure}[htbp]
\centering
\includegraphics[scale=0.5]{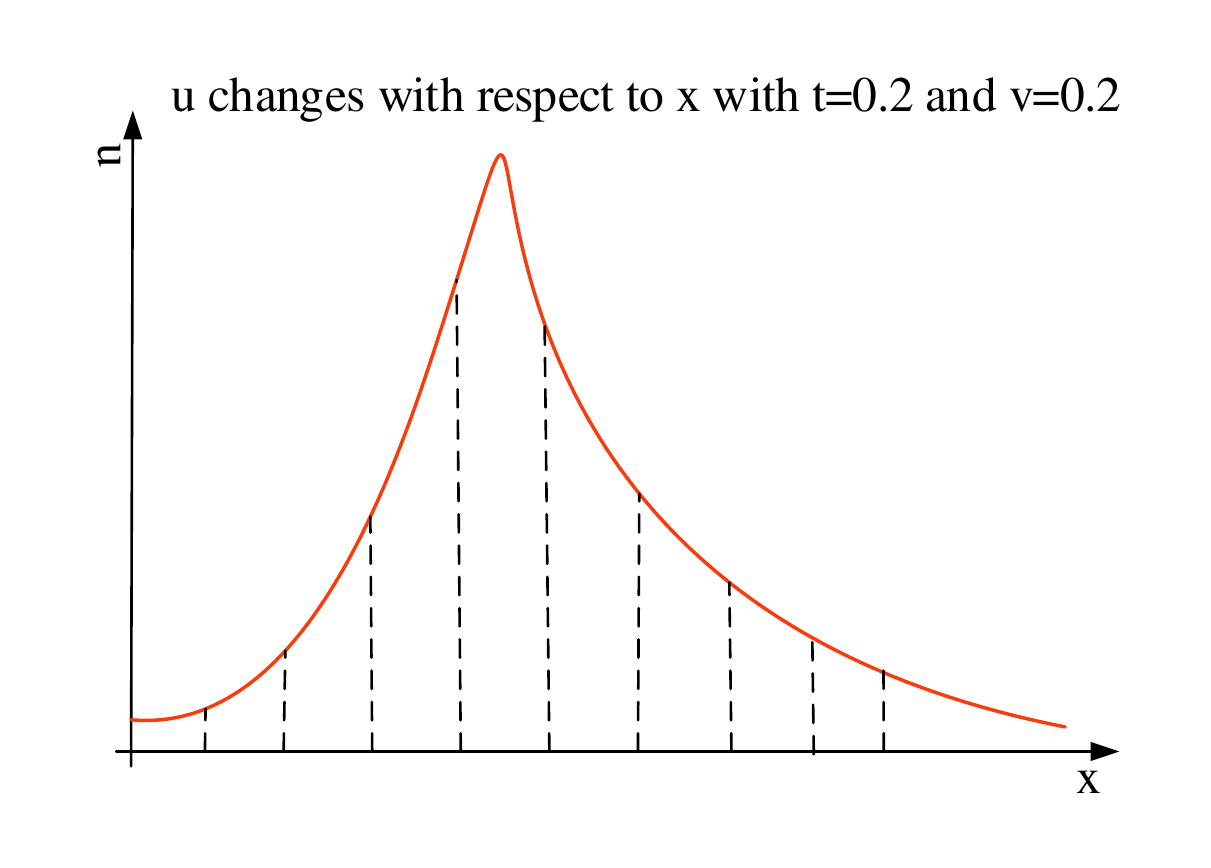}
\caption{The uniform sampling in Burgers' equation. $x$ denotes the displacement and $u$ denotes the velocity. }\label{fig_problem}
\centering
\end{figure}

In the field of aerodynamic data predictions, the quality of datasets is one of the key factors that influence the accuracy of prediction models \cite{yang2018dynamic}. For example, in Burgers' equation, uniform sampling results in limited number of data with high-speed, see Fig.\ref{fig_problem}. Limited high-speed data are submerged in lots of low-speed data, which leads to the inability for neural networks to predict high-speed aerodynamic data. This phenomenon, more common in unsteady flow fields \cite{thormann2017efficient}, increases the difficulty of modeling and also reduce the accuracy of models.

To improve the quality of datasets, data resampling technology is widely used in this field \cite{benyamen2020effects}. The aim of data resampling is to produce data conforming to the law of aerodynamic variations for the parts of flow fields with insufficient data to alleviate the defects caused by poor quality of datasets. Generally, the studies of data resampling can be divided into two categories: data recalculation and data generation.

Data recalculation is a traditional method that relies on computational fluid dynamics (CFD) solvers \cite{li2020spectral}. In aerodynamics, the CFD solvers usually simulate Navier-Stokes equations to express the variations of air flows \cite{zhang2020coupled}. However, even optimal CFD solvers, e.g. SU2 \cite{vitale2020multistage} or Fluent \cite{shah2020comparative}, take weeks to complete a high-dimensional aerodynamic calculation, which affects the instantaneity of predictions about fluid dynamics \cite{he2020cfd}.

Data generation, usually based on neural networks, achieves great success both in the field of machine learning and aerodynamics \cite{hu2020neural}. This method adopts generative models to generate plentiful aerodynamic data. Among these generative models, the generative adversarial network (GAN) \cite{gui2020review}, the most promising deep learning model in recent years, is pervasively used in aerodynamic data generation \cite{hu2020neural}. The recursive CNNs-based GAN in unsteady flow predictions \cite{lee2019data}, the RNNs-based GAN in the analyses of the temporal continuity \cite{kim2020deep}, the tempoGAN for high-resolution flow field image generations \cite{xie2018tempogan} and the RBF-GAN/RBFC-GAN for nonlinear sparse flow field data generation \cite{hu2020flow} are outstanding models.

Recently, in the field of natural language processing, the mixtures of experts have been applied to analyze complex sentences, which have achieved great success \cite{zhang2019learning}. The idea of mixtures of experts is that different areas in a neural network are responsible for different parts of speech, i.e. a small area in a neural network is dedicated to processing a specific part, not the entire sentence. This mechanism provides a solution for eliminating the need for data resampling in aerodynamic data modeling, owing to that parts of data in flow fields can be processed by a dedicated area in neural networks-based aerodynamic models.

Motivated by this idea, we propose the multi-task learning (MTL) scheme which consists of two parts: task allocation and aerodynamic characteristics learning. The task allocation divides a complete aerodynamic task (dataset) into multiple subtasks (subsets) according to specific methods. In this paper, the data partition or K-means method \cite{ball1967clustering} is employed to achieve this goal. Data partition, a vertical task allocation, divides a whole dataset into several subsets based on a certain dimension in the dataset. While K-means, a horizontal task allocation, clusters the samples in a dataset into multiple subsets. The aerodynamic characteristics learning learns the aerodynamic variations contained in each subsets at the same time, and outputs the corresponding predictions based on given inputs. In this paper, the ClusterNet is applied to implement the aerodynamic characteristics learning \cite{white2020fast}. Two experiments with poor quality of datasets are conducted to verify the feasibility of the MTL. Compared with fully connected networks (FCNs) and GANs, the MTL can accurately predict the velocity $u$, the coefficient of pressure $C_{P}$ and the friction $F_{x}$ in the entire flow field from datasets with poor quality. The results reflects than the MTL is a dataset quality-adaptive learning scheme. Besides, the MTL explains in detail how the ClusterNet automatically recognizes regions in datasets, which was not explained in the original paper \cite{white2020fast}.

To summarize, the contributions of our work are:

a) we propose the MTL in aerodynamic data modeling;

b) our work eliminates the data resampling, while dealing with poor quality aerodynamic datasets;

c) our work, for the first time, explains in detail how the ClusterNet automatically recognizes regions in datasets.

The structure of the remainder of this paper is as follows. Section II introduces the research status of K-means clustering algorithm and the ClusterNet in the field of aerodynamic response predictions. In Section III, the details of MTL are elaborated. In Section IV, two typical datasets from different scenes are applied to validate the effectiveness of the proposed scheme. The conclusion of our work is shown in Section V.

\section{Previous Work}

The MTL consists of two parts: task allocation and the aerodynamic characteristics learning. As the task allocation mainly uses the K-means method, and the aerodynamic characteristics learning mainly relies on the clusterNet model, in this section, we introduce the previous work of the K-means method and the ClusterNet model.

\subsection{K-means Method}

Ball presented the K-means method in 1967 and since then it has become the most popular of the clustering algorithms \cite{ball1967clustering}. In aerodynamics, K-means method is widely used to data clustering and parameters estimation.

Organizing aerodynamic data into multiple sensible groups is a foundation of MTL. Usually, it is difficult to divide a whole aerodynamic dataset into several subsets even for experts, owing to the complex variations of air flows. K-means, as a fast and flexible unsupervised learning method, provides a viable solution \cite{jain2010data}. \cite{taguchi2020experimental} adopted k-means to obtain representative points on the upper and lower surface of a airfoil. \cite{bamberger2016aerodynamic} applied K-means to optimize data distribution to cover the input space as uniform as possible. \cite{edmunds2016enhanced} replaces the hierarchical clustering algorithm with a K-means that is suitable for large unstructured grids.

K-means algorithms are also used to estimate the hyper parameters. For example, the centers of a radial basis function can be determined by clustering the input data \cite{sanwale2018aerodynamic}. Besides, the parameters (weights and biases) of other models can also be initialized with k-menas \cite{singh2017aerodynamic,li2019data}.

Different from the above studies, we employ K-means to clustering aerodynamic data into different independent subsets to provide learning labels for MTL.

\subsection{ClusterNet Model}

For general neural networks, each neuron will be activated both in training and testing process, which is difficult to adapt to poor quality learning samples. Although, large-scale neural networks seems to be a feasible approach \cite{hamerly2019large,tops2017large}, the computational costs increase dramatically. The reason for this is that all neurons are activated for each sample, which is clearly different from the way the brain processes informations from real world \cite{zhou2016learning}.

The origin of ClusterNet can be traced back to mixtures of experts (i.e. dedicated neural networks) in natural language processing \cite{jacobs1991adaptive}. The idea of mixtures of experts is that different experts are responsible for analyzing different parts of speeches. The final results are obtained by linearly weighting results of all experts \cite{shazeer2017outrageously}. The experts can be implemented by different models, such as SVMs \cite{rajaei2013human}, gaussian processes \cite{deisenroth2015distributed}, neural networks \cite{zhang2019learning}, etc. After 2019, the mechanism of partial activation of neurons is developed in the form of clusterNet in aerodynamics.

The clusterNet is a novel neural network in that different clusters automatically identify different regions in datasets. \cite{white2020fast} proposed the ClusterNet in 2020, which is used to predict the velocity of air flows. Compared with general neural networks, ClusterNet reduced the errors of predicted velocity. \cite{zhang2020learning} adopted a clusterNet-based physics-informed model to predict future trajectories of the swarm by approximating the nonlinear dynamics of the swarm model, which is more stable than other models in nonlinear ordinary differential equations systems.

However, how the model automatically identify different regions in datasets and why the ClusterNet is better than others is not analyzed in detail. Therefore, we combine task allocation and the ClusterNet to form MTL, which effectively explains the mechanism of the ClusterNet.

\section{Methodology}

\subsection{Overview}

\begin{figure}[htbp]
\centering
\subfigure[]{
\includegraphics[scale=0.3]{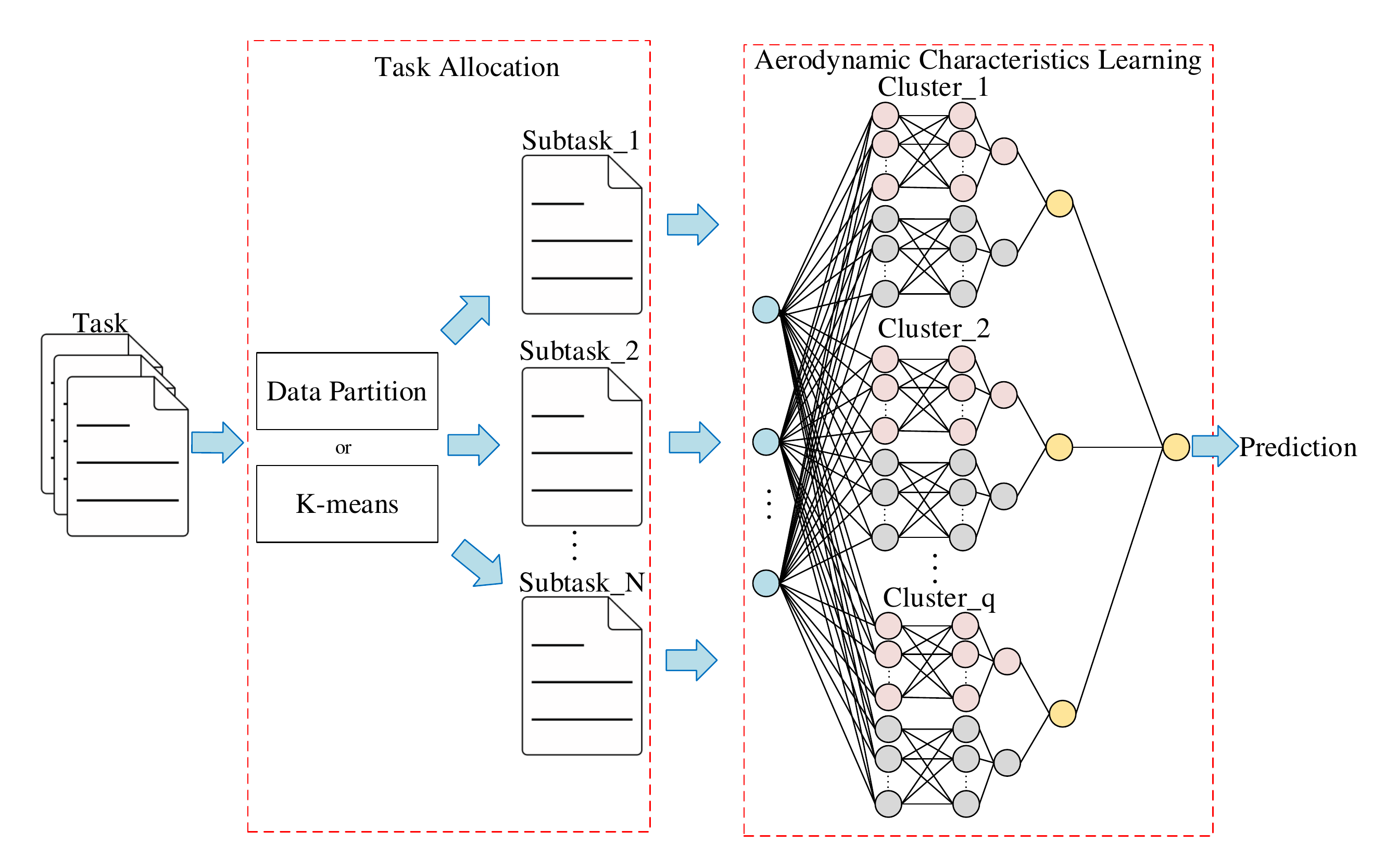}
}
\quad
\subfigure[]{
\includegraphics[scale=0.41]{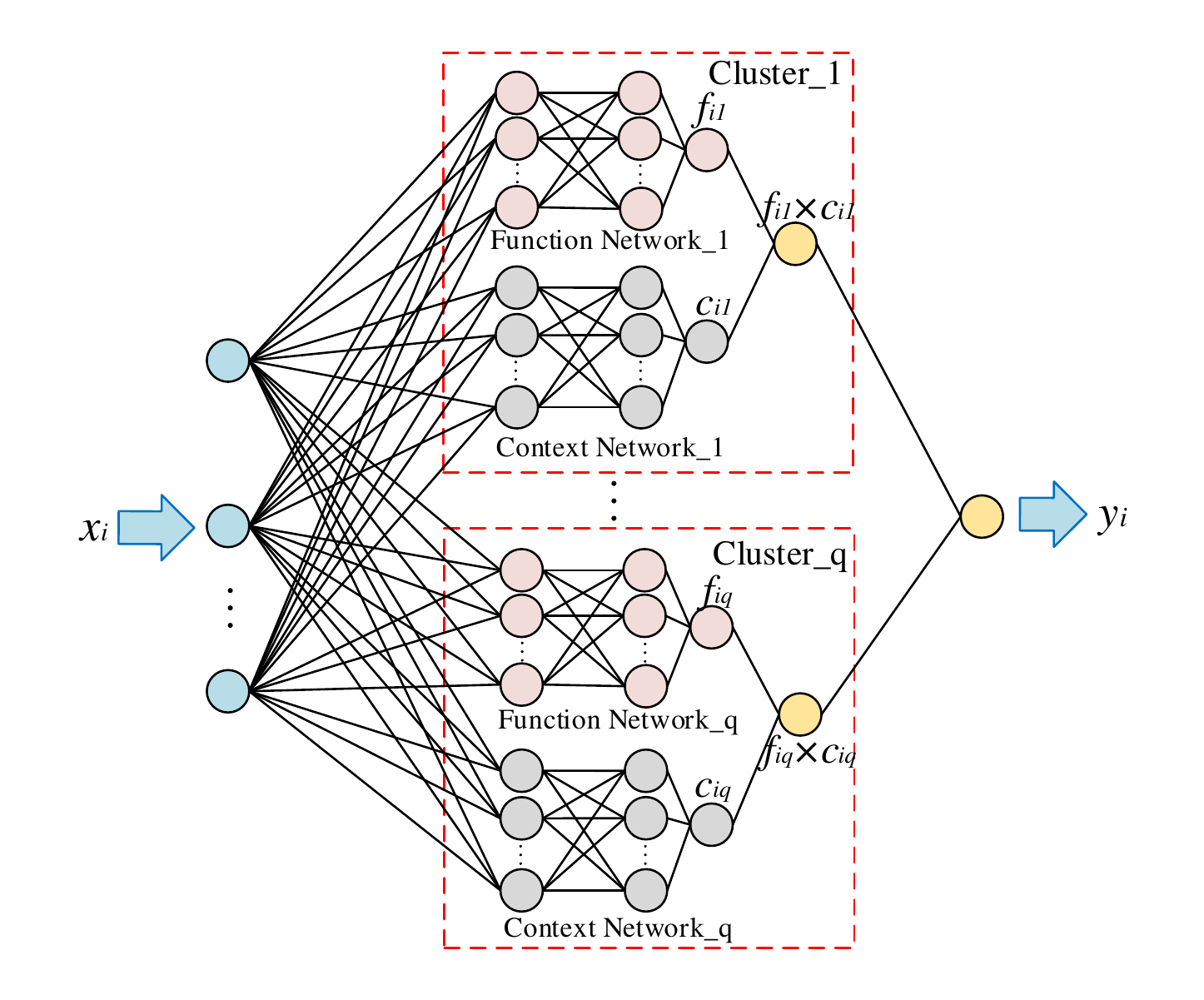}
}
\caption{(a) The scheme of MTL in aerodynamic data modeling. (b) The structure of a ClusterNet, in which the pink neurons denote the function network and the gray neurons denote the context network.}\label{fig_structure}
\end{figure}

In this section, we introduce the MTL which consists of task allocation and aerodynamic characteristics learning, see Fig.\ref{fig_structure} (a). As for the task allocation, we use data partition or K-means method to divide a whole learning task into multiple independent subtasks according to specific rules. As for the ClusterNet, we use both a classification loss function and a regression loss function to learn these subtasks at the same time to achieve the purpose of partial activation.

\subsection{Task Allocation}

Task allocation divides a given dataset into multiple independent subsets. Given a dataset $\mathbf{D}  \subseteq \mathbf{R}^{n\times m}$, the goal of task allocation is to divide $\mathbf{D}$ into $k$ independent subsets $\mathbf{\hat{D}}=\{\mathbf{\hat{D}}_{z}|\cap_{z=1}^{k}\mathbf{\hat{D}}_{z}=\emptyset\}$. In this paper, we provide two alternative methods: data partition and K-means method.

\subsubsection{Data Partition}

We analyze $\mathbf{D}$ from the perspective of column vector, i.e. $\mathbf{D}=\{\mathbf{d}_{1},\mathbf{d}_{2},\cdots,\mathbf{d}_{m}\},$ where $\mathbf{d}_{i}$ denotes the $i$th dimension of $\mathbf{D}$. Assuming that for any $v \in \mathbf{d}_{i}$ satisfies $a \leq v  \textless b$ , then $\hat{v}$ satisfies:

\begin{equation}
\begin{cases}
a+\sum_{j=0}^{z-1}\lambda_{ij}\leq \hat{v} \textless a+\sum_{j=0}^{z}\lambda_{ij},z=0,1,2,\cdots,k-1\\
a+\sum_{j=0}^{z-1}\lambda_{ij} = b\\
\hat{v} \in \hat{\mathbf{d}}_{iz}\\
\hat{\mathbf{d}}_{iz} \subseteq \mathbf{\hat{D}}_{z}, z=1,2, \cdots ,k
\end{cases}
\end{equation}
where $\mathbf{\hat{D}}_{z}$ denotes the $z$th subset, $\mathbf{\hat{d}}_{iz}$ denotes the $i$th dimension of the $z$th subsets $\mathbf{\hat{D}}_{z}$, $\hat{v}$ denotes the elements of $\mathbf{\hat{d}}_{iz}$ and $\lambda_{ij}$ denotes the change length of the $i$th dimension data in the $j$th subset.

\subsubsection{K-means method}

We analyze $\mathbf{D}$ from the perspective of row vectors, i.e. $\mathbf{D}=\{\mathbf{x}_{1},\mathbf{x}_{2},\cdots,\mathbf{x}_{n}\}^\mathsf{T}$ where $n$ denotes the number of data. The goal of K-means method is to cluster all the data in $\mathbf{D}$ into a set of $k$ clusters (subsets), i.e. $\mathbf{\hat{D}}$. K-means method finds a clustering result to make the following formula reach its minimum value:

\begin{equation}
min(L) = \sum_{z=1}^{k}\sum_{x_{r}\in \mathbf{\hat{D}}_{z}}||x_{r}-\mu_{z}||^{2}
\end{equation}
where $x_{r}$ denotes the data in $\hat{D}_{z}$, $\mu_{z}$ denotes the center (mean) of $\hat{D}_{z}$. The algorithm obtains good clustering results by continuously adjusting the center $\mu_{z}$.

K-means method is a greedy algorithm, which starts with an initial partition with $K$ clusters. Usually, this method is considered to converge to a local minimum, while \cite{meilua2006uniqueness} showed that it can converge to the global minimum with a large probability. The details of K-means method can be found in \cite{jain1988algorithms}.

\subsection{ClusterNets}

The ClusterNet is unique in that it can process multiple subtasks simultaneously, which is due to the special structure and the training method.

\subsubsection{The Sturcture}

As shown in Fig.\ref{fig_structure} (b), a ClusterNet consists of multiple clusters, each of which is responsible for a specific subtask, i.e. $q=k$, where $q$ denotes the number of clusters. A cluster is composed of two parts: a function network and a context network. The function network learns the conditional probability $P(X|Y)$ to predict the corresponding value for given inputs, i.e. the function network is a regression model. While the context network learns the nonlinear relationship between the inputs and its category labels (from task allocation) to identify the subtask the cluster should handle, i.e. the context network is a classification model.

The output of a ClusterNet can be written as:

\begin{equation}
y_{i}=\mathop{\sum_{j=1}^{q}f_{ij}\times c_{ij}}
\end{equation}
where $f_{ij}$ and $c_{ij}$ denote the output values of the function network and the context network in the $j$th cluster corresponding to the $i$th input, respectively.

\subsubsection{The Training Method}
In a ClusterNet, the function network and the context network are trained alternately to achieve both regression and classification simultaneously.

As for the regression, the loss function of function networks
is:

\begin{equation}
\label{equ_function_loss}
L_{f}=\frac{1}{N}\sum_{i=1}^{N}(y_{i}-\hat{y}_{i})^{2}
\end{equation}
where $N$ denotes the number of data in the training set, $y_{i}$ denotes the predicted value, and $\hat{y_{i}}$ denotes the real value in the training set. Formula (\ref{equ_function_loss}) dose not involve any information of context networks, therefore only the parameters of function networks in all clusters will be updated at this step.

As for the classification, the loss function of context networks
is :

\begin{equation}
\label{equ_context_loss}
L_{c}=\frac{1}{N}\sum_{i=1}^{N}L_{i}=\frac{1}{N}\sum_{i=1}^{N}[-\sum_{j=1}^{q}c_{ij}log(p_{ij})]
\end{equation}
where $c_{ij}$ denotes the probability that the $i$th input is classified into the $j$th category by the $j$th context network, and $p_{ij}$ denotes the true category the $i$th input belongs to. Formula (\ref{equ_context_loss}) does not involve any information of function networks, hence only the parameters of context networks in all clusters will be updated at this step.

In conclusion, the training of function networks is only related to $L_{f}$, while the training of context networks is only related to $L_{c}$. The updating process of parameters in a ClusterNet is shown as:
\begin{equation}
\label{equ_update}
\begin{cases}
\mathbf{W}_{fj}=\mathbf{W}_{fj}+\eta\frac{\partial L_{f}}{\partial \mathbf{W}_{fj}}\\
\mathbf{b}_{fj}=\mathbf{b}_{fj}+\eta\frac{\partial L_{f}}{\partial \mathbf{b}_{fj}}\\
\mathbf{W}_{cj}=\mathbf{W}_{cj}+\eta\frac{\partial L_{c}}{\partial \mathbf{W}_{cj}}\\
\mathbf{b}_{cj}=\mathbf{b}_{cj}+\eta\frac{\partial L_{c}}{\partial \mathbf{b}_{cj}}\\
\end{cases}
\end{equation}
where $\mathbf{W}_{fj}$ and $\mathbf{b}_{fj}$ denotes the weight matrix and bias vector of the function network in the $j$th cluster, respectively, while $\mathbf{W}_{cj}$ and $\mathbf{b}_{cj}$ denotes the weight matrix and bias vector of the context network in the $j$th cluster, respectively.

\section{Experimental Results}

\subsection{Dataset}

\begin{figure}[htbp]
\centering
\subfigure[]{
\includegraphics[scale=0.13]{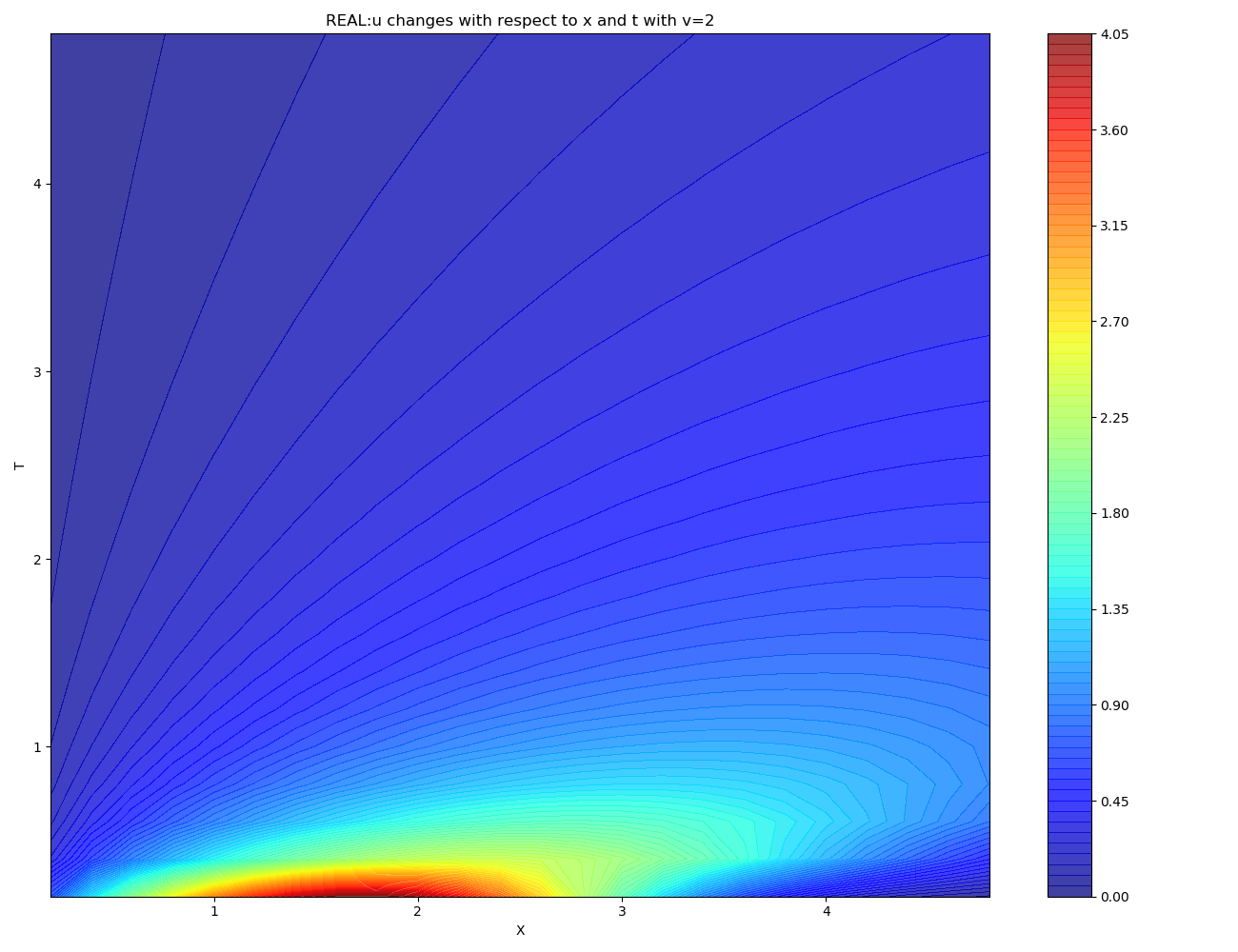}
}
\quad
\subfigure[]{
\includegraphics[scale=0.13]{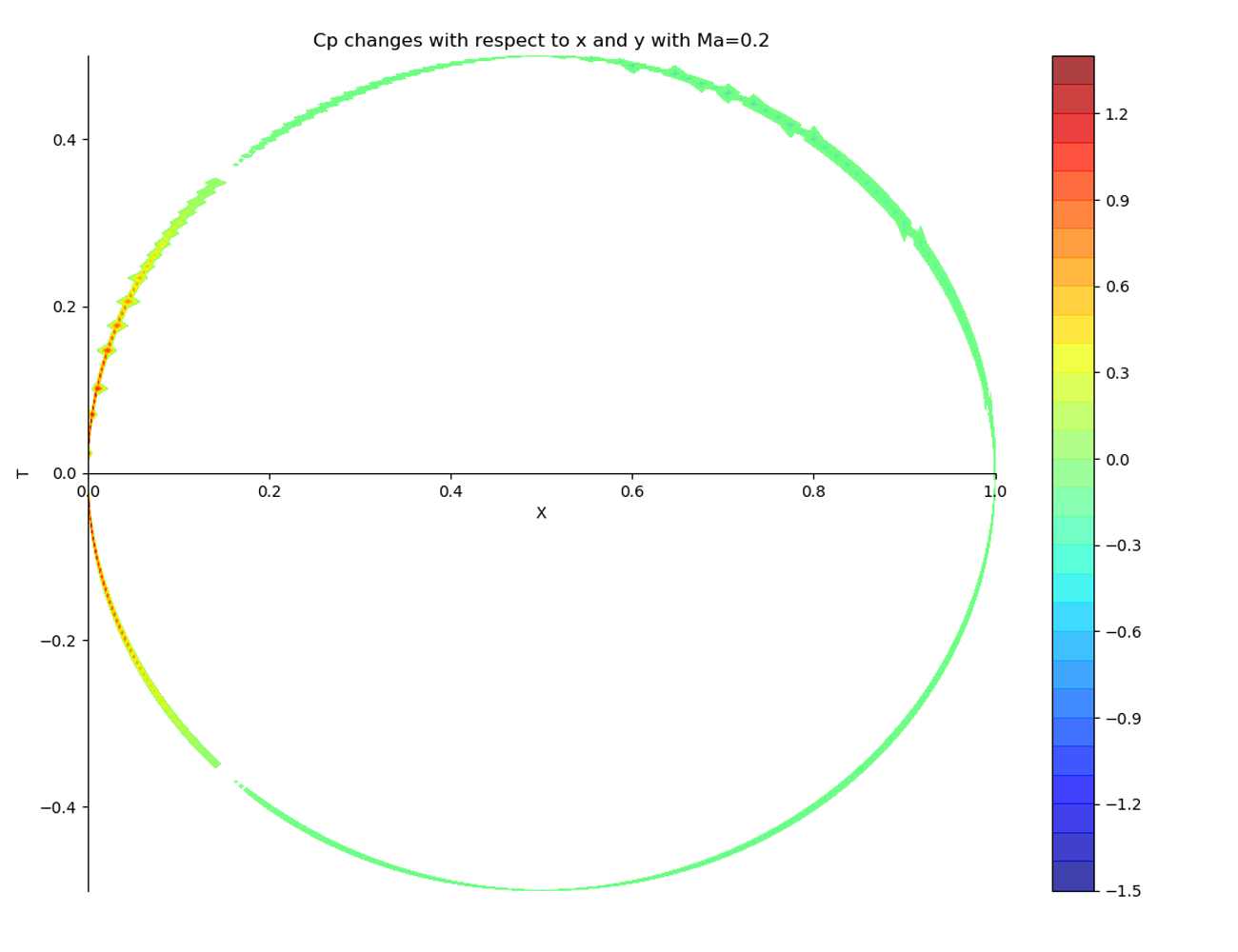}
}
\caption{Subgraph (a) denotes the visualization of Burgers' dataset and (b) denotes the visualization of cylindrical laminar dataset.}\label{fig_visualization_of_dataset}
\end{figure}

To validate the dataset quality-adaptivity of the MTL, two different datasets are employed, namely the Burgers' dataset and the cylindrical laminar dataset.

\subsubsection{Burgers' Dataset}

The Burgers' equation is an one-dimensional partial differential equation that expresses the movement of a shockwave across a tube:

\[
\ensuremath{\frac{\partial u}{\partial t}+u\frac{\partial u}{\partial x}=v\frac{\partial^{2}u}{\partial x^{2}}}
\]
where $u$, the output, denotes the velocity of the shockwave, $t$, $x$, and $v$, the inputs, denotes the time, the displacement and the coefficient of viscosity, respectively. The variation range of every input is from 0.2 to 4.8, and the step is 0.2. Consequently, this dataset contains 13824 samples. Fig.\ref{fig_visualization_of_dataset} (a) illustrates the visualization of whole Burgers' dataset. The red area in the lower left corner depicts the high-speed flow field data. We can see that the low-speed flow field data are plentiful, while the number of high-speed flow field data is limited, which can be used to verify the dataset quality-adaptivity of MTL.

\subsubsection{Cylindrical Laminar Dataset}

\begin{table}[tbh]
\centering
\begin{tabular}{cccc}
\hline
variables & $x$ & $y$ & $Ma$\tabularnewline
\hline
significance & x-coordinate & y-coordinate & mach number\tabularnewline
start & 0.1 & -0.5 & 0.1\tabularnewline
end & 1 & 0.5 & 0.24\tabularnewline
step & 0.005 & 0.0078 & 0.01\tabularnewline
\hline
\end{tabular}

\caption{The data format and variation range of input parameters in cylindrical laminar dataset.}\label{tab_NS_input_data_format}
\centering
\end{table}

\begin{table}[tbh]
\centering
\begin{tabular}{ccccc}
\hline
variables & $P$  & $C_{p}$ & $F_{x}$ & $F_{y}$\tabularnewline
\hline
significance & pressure & coefficient of pressure & friction in x-coordinate & friction in y-coordinate\tabularnewline
\hline
\end{tabular}

\caption{The data format of output parameters in cylindrical laminar dataset.}
\label{tab_NS_output_data_format}
\centering
\end{table}

The cylindrical laminar is a two-dimensional application of the Navier-Stokes equation that simulates the pressure change on the surface of a cylinder while a flow passes a cylinder. The Navier-Stokes equation is as follows:

\[
\ensuremath{\rho\left[\frac{\partial V}{\partial t}+(V.\nabla)V\right]=-\nabla P+\rho g+v\nabla^{2}V}
\]
where $P$, $V$, $t$, $\rho$ , $v$ denote the pressure, velocity, time, density and coefficient of viscosity, respectively. In this dataset, we used SU2 to calculate 6000 sample points. The format and the variation range of the input parameters are shown in Tab.\ref{tab_NS_input_data_format}. The format of the output parameters is shown in Tab.\ref{tab_NS_output_data_format}.

The visualization of cylindrical laminar dataset is shown in Fig.\ref{fig_visualization_of_dataset}(b) that reflects the same feature as Fig.\ref{fig_visualization_of_dataset}(a): the number of data with large $C_{p}$ is limited.

\subsection{Experiment Results}

We compare several approaches including FCN, GAN and MTL with different structures. The learning rate is 0.0001, with the batch size 128, and the number of iterations 2000 within all models. All these approaches compared in this paper are implemented based on tensorflow framework \cite{pang2020deep}. All programs run on four Tesla K80 GPUs. The whole dataset is divided into three subsets: a training set, a validation set and a test set with a ratio of 8:1:1. The errors of these models are evaluated by mean squared error (MSE) and mean absolute error (MAE):

\begin{equation}
\nonumber
\begin{cases}
MSE=\frac{1}{n}\sum_{i=1}^{n}\left(y_{i}-\hat{y_{i}}\right)^{2}\\
MAE=\frac{1}{n}\sum_{i=1}^{n}|y_{i}-\hat{y_{i}}|
\end{cases}
\end{equation}
where $y_{i}$ denotes the predicted value, $\hat{y_{i}}$ denotes the real value for the same inputs.

\subsubsection{Burgers' Experiment}

\begin{table}[tbh]
\centering
\begin{tabular}{cccc}
\hline
Method & structure & MSE & MAE\tabularnewline
\hline
FCN\_1 & 3{*}32 & $1.64\times10^{-4}$ & $7.71\times10^{-3}$\tabularnewline
FCN\_2 & 3{*}64 & $1.85\times10^{-4}$ & $8.82\times10^{-3}$\tabularnewline
cGAN\_1 & G(62,1{*}64,4)D(4,1{*}64,1) & $2.63\times10^{-4}$ & $1.29\times10^{-2}$\tabularnewline
cGAN\_2 & G(62,3{*}64,4)D(4,3{*}64,1) & diverges & diverges\tabularnewline
MTL\_k-means & 4;3{*}64;1{*}5 & $1.79\times10^{-4}$ & $8.11\times10^{-3}$\tabularnewline
MTL\_v & 4;3{*}64;1{*}5 & $1.78\times10^{-4}$ & $8.91\times10^{-3}$\tabularnewline
MTL\_x & 4;3{*}64;1{*}5 & $9.86\times10^{-5}$ & $6.70\times10^{-3}$\tabularnewline
\hline
\centering
\end{tabular}

\caption{The Burgers' experiment results. In the column related to structure, \textquotedblleft 3{*}32\textquotedblright{} denotes that the FCN has 3 hidden layers, each of which has 32 neurons. ``G(62,1{*}64,4)'' indicates that generator of the GAN has 3 layers, the number of neurons in the input layer, the hidden layer and the output layer is 62, 64 and 4, respectively. ``D(4,1{*}64,1)'' indicates that the discriminator of the GAN has 3 layers, 4, 64 and 1 denotes the number of neurons in the input layer, the hidden layer and the output layer, respectively. ``4;3{*}64; 1{*}5'' means that the ClusterNet consists of 4 clusters. The functional network has 3 hidden layers, each of which has 64 nodes. The context network has 1 hidden layer, each of which has 5 neurons. MTL\_k-means denotes the multi-task learning that uses k-means for subtask division. MTL\_v denotes the multi-task learning whose subtasks are divided by $v$. Similarly, MTL\_x denotes that the subtasks are divided by $x$.}
\label{tab_burgers_models_compare}
\end{table}

The results of Burgers' experiment are shown in Tab.\ref{tab_burgers_models_compare}. We can learn that MTL\_x is the optimal approach in term of MSE and MAE. Besides, the rest two MTLs are similar to FCNs, and the cGANs are the worst among them all.

\begin{figure}[htbp]
\centering
\subfigure[]{
\includegraphics[scale=0.14]{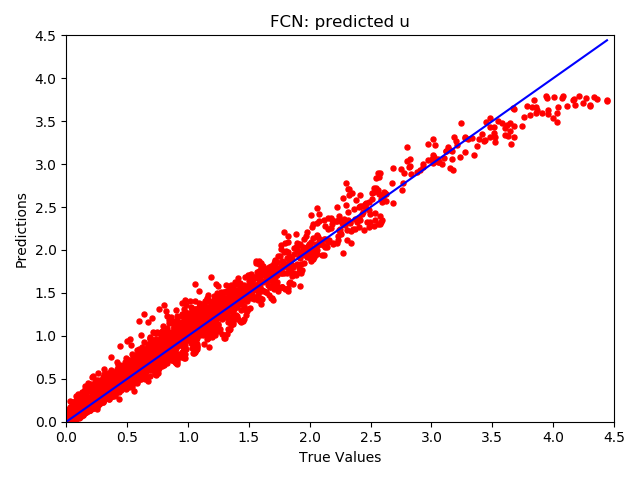}
}
\quad
\subfigure[]{
\includegraphics[scale=0.14]{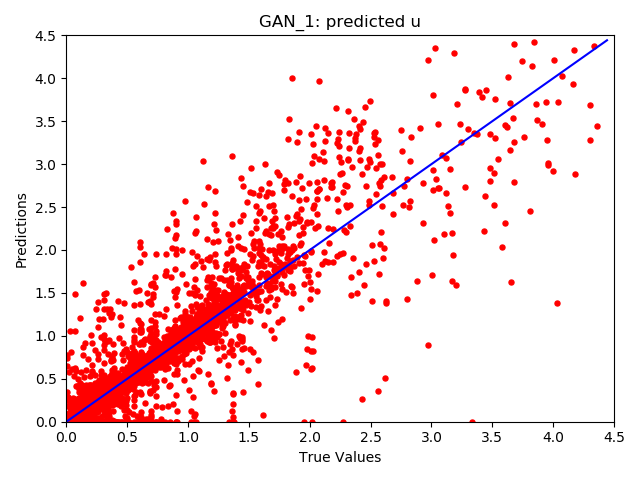}
}
\quad
\subfigure[]{
\includegraphics[scale=0.14]{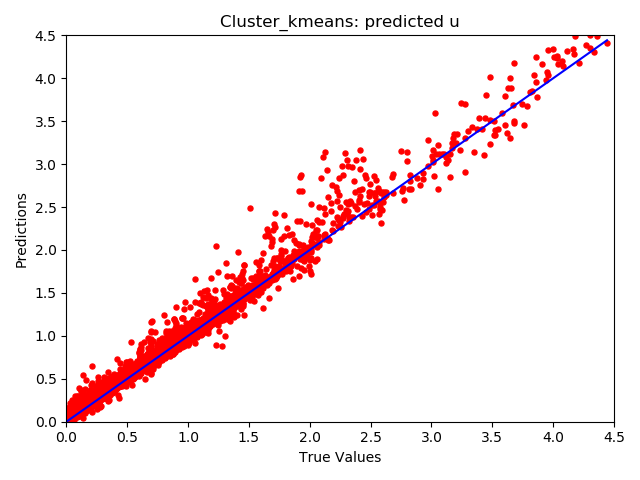}
}
\quad
\subfigure[]{
\includegraphics[scale=0.14]{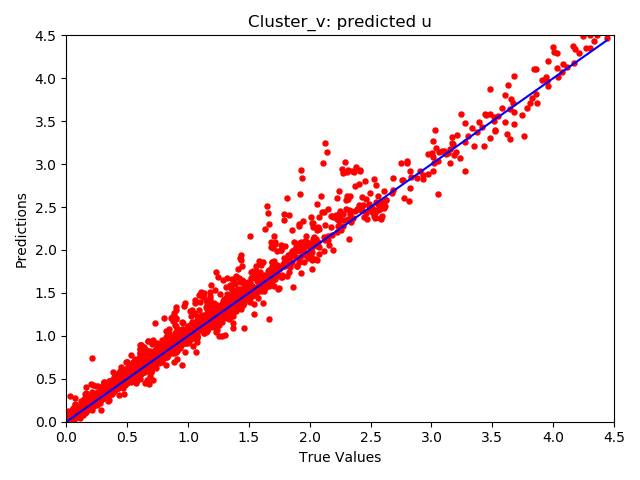}
}
\quad
\subfigure[]{
\includegraphics[scale=0.14]{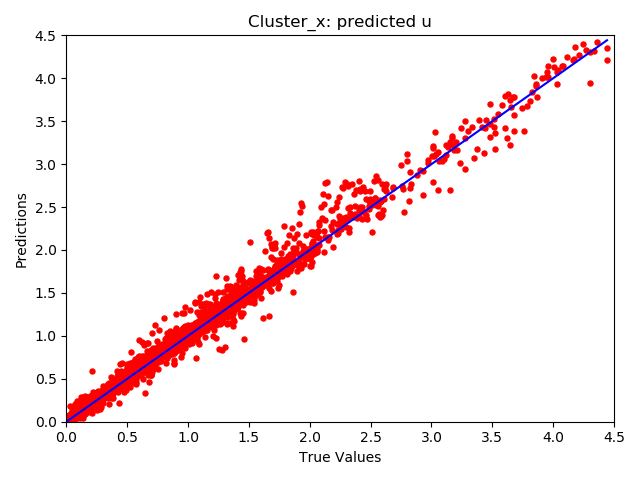}
}

\caption{The $u$ predicted by FCN\_1 (a), cGAN\_1 (b), MTL\_k-means (c), MTL\_v (d) and MTL\_x (e).}\label{fig_burgers_predicted_u}
\centering
\end{figure}

The visualization results of velocity $u$ predicted by the above approaches are shown in Fig.\ref{fig_burgers_predicted_u}. Obviously, the FCN can not predicted $u$ precisely where $u>3.5$. What's worse, the cGAN could not achieve accurate prediction of $u$ in the entire flow field. Therefore, we can get a conclusion that the velocity $u$ predicted by MTLs are much more accurate than FCNs and cGANs, i.e. the MTL is a dataset quality-adaptive learning scheme.

\begin{figure}[htbp]
\centering
\subfigure[]{
\includegraphics[scale=0.12]{Burgers_real}
}
\quad
\subfigure[]{
\includegraphics[scale=0.12]{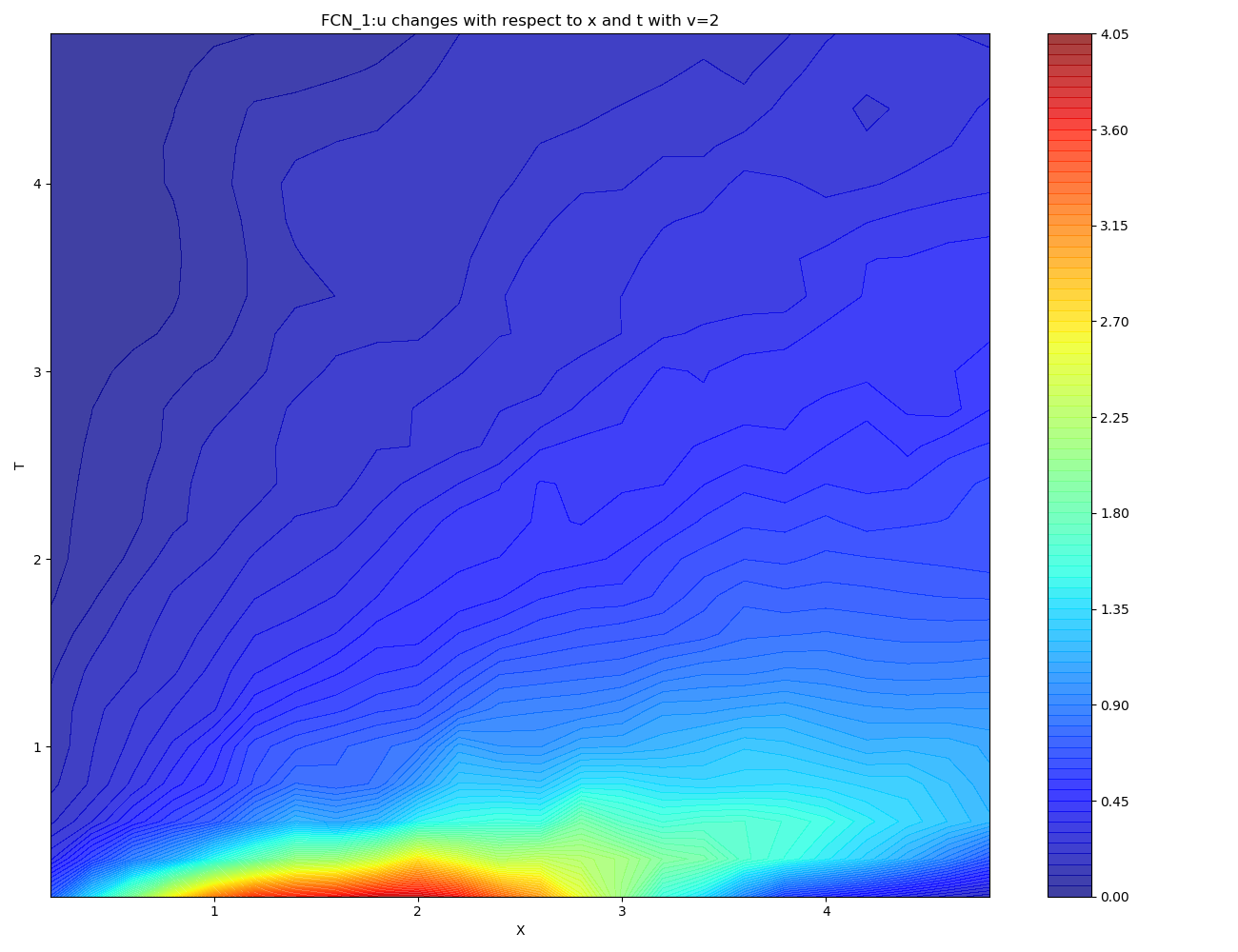}
}
\quad
\subfigure[]{
\includegraphics[scale=0.12]{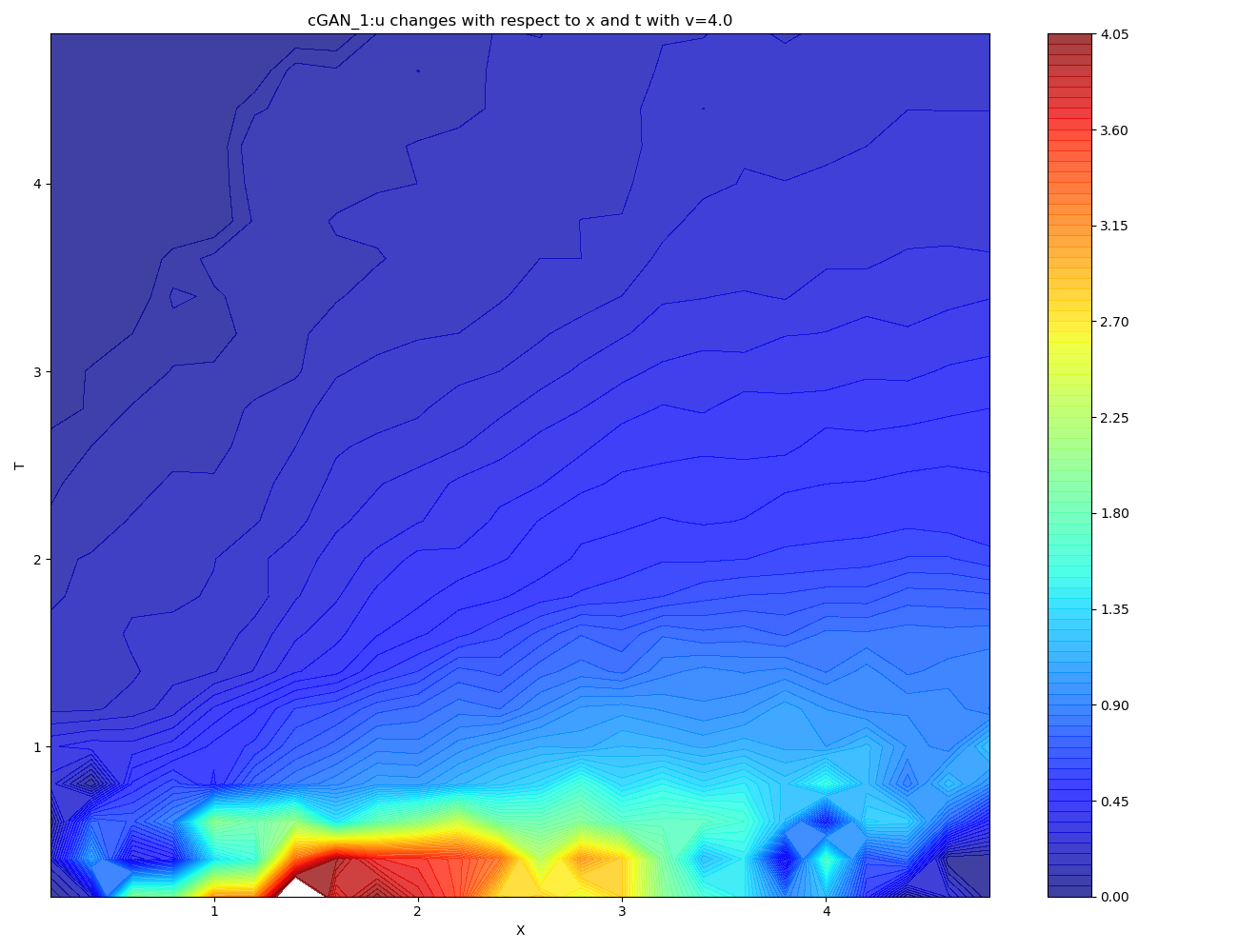}
}
\quad
\subfigure[]{
\includegraphics[scale=0.12]{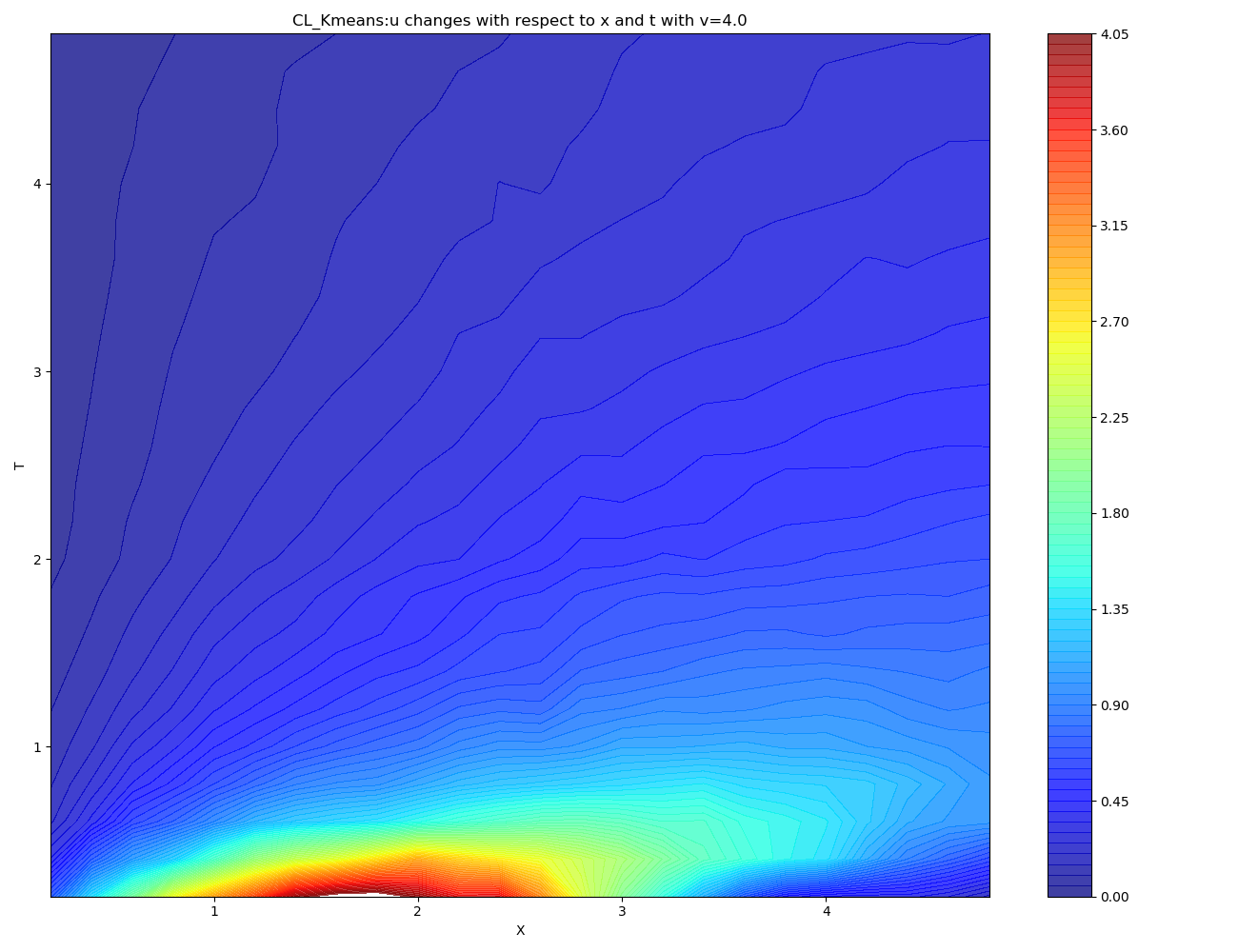}
}
\quad
\subfigure[]{
\includegraphics[scale=0.12]{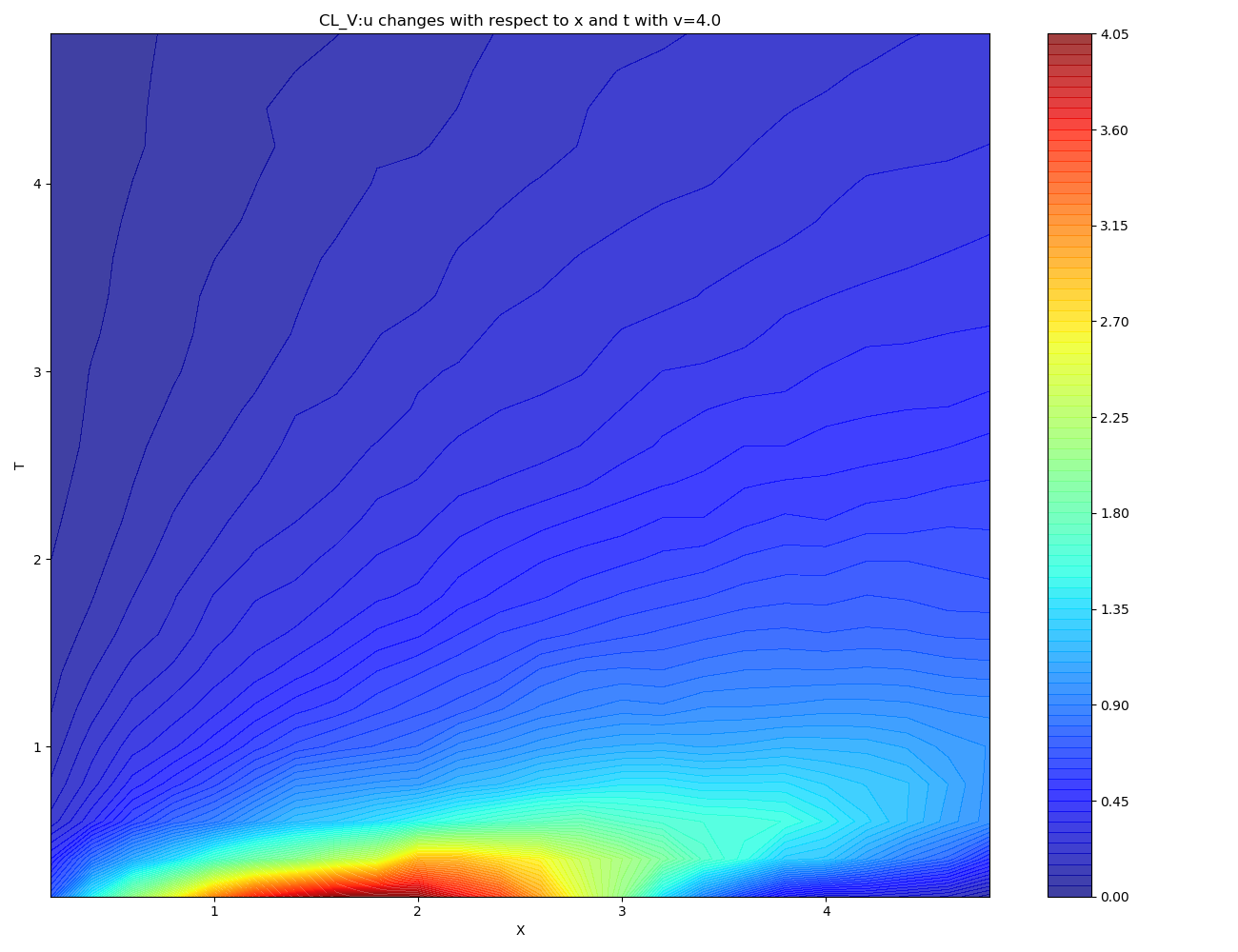}
}
\quad
\subfigure[]{
\includegraphics[scale=0.1]{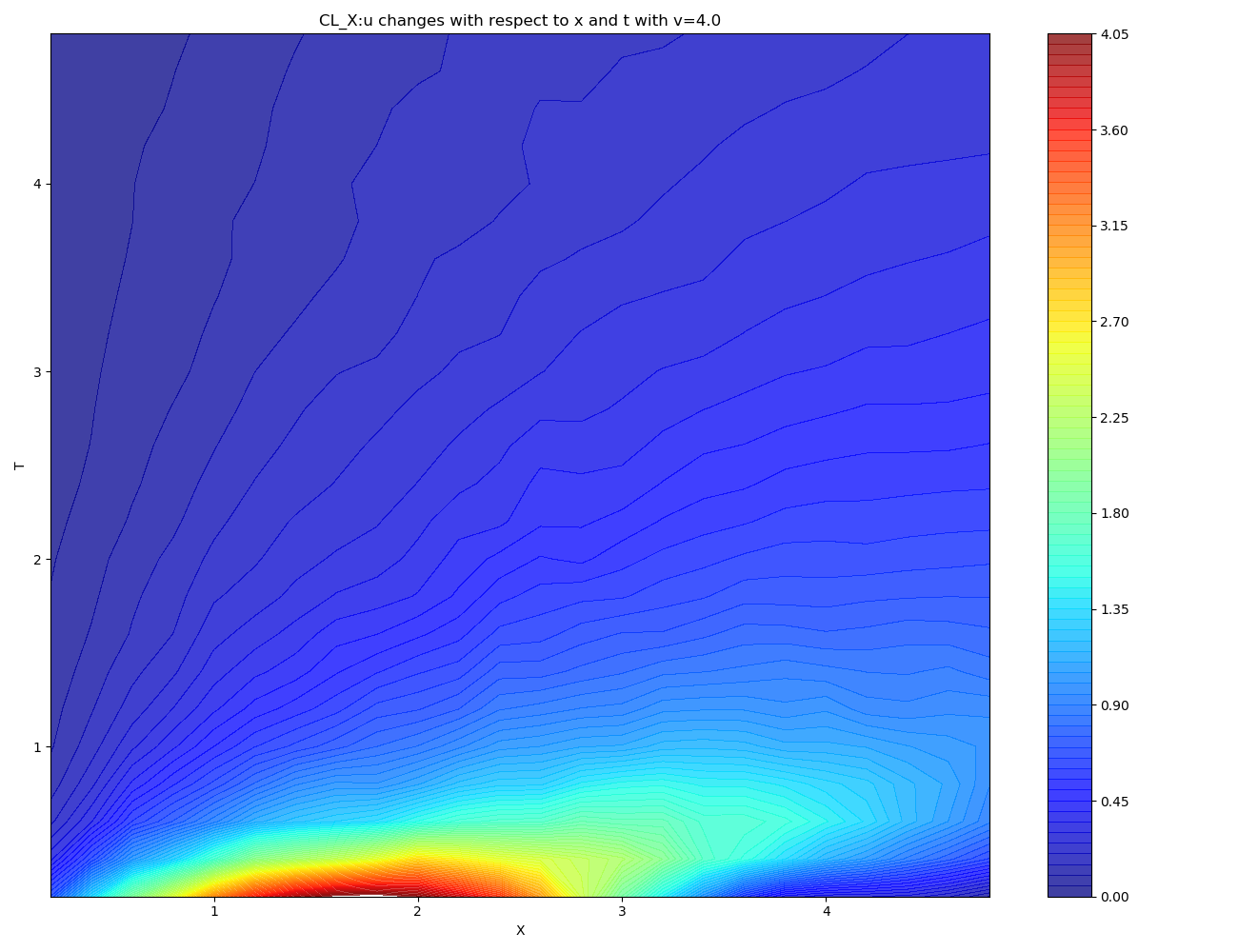}
}
\caption{The $u$ predicted by CFD (a), FCN\_1 (b), cGAN\_1 (c), MTL\_k-means (d), MTL\_v (e) and MTL\_x (f).}\label{fig_burgers_visualization}
\centering
\end{figure}

Fig.\ref{fig_burgers_visualization} also reflects the same results. The red part in subgraph (b) is lighter than others, which means than the value of predicted $u$ is smaller than that of other approaches. Besides, the contour line in subgraph (b) and (c) are not smooth, which means that the predicted $u$ of the FCN and the cGAN are very different from the real one. It is worth noting that subgraph (d), (e) and (f), i.e. the three MTLs, are much more close to subgraph (a).

\begin{figure}[htbp]
\centering
\subfigure[]{
\includegraphics[scale=0.2]{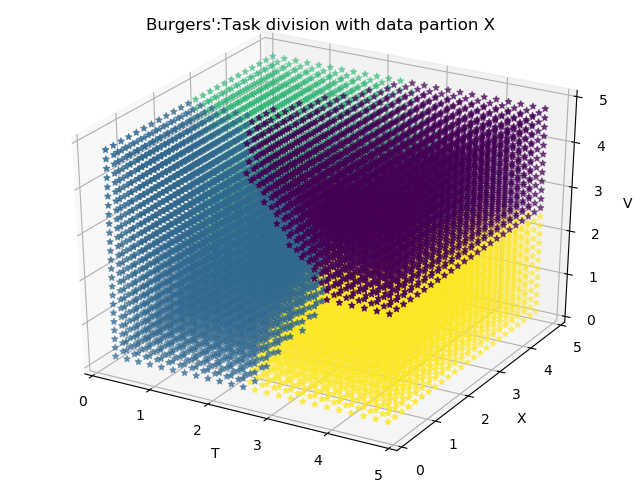}
}
\quad
\subfigure[]{
\includegraphics[scale=0.2]{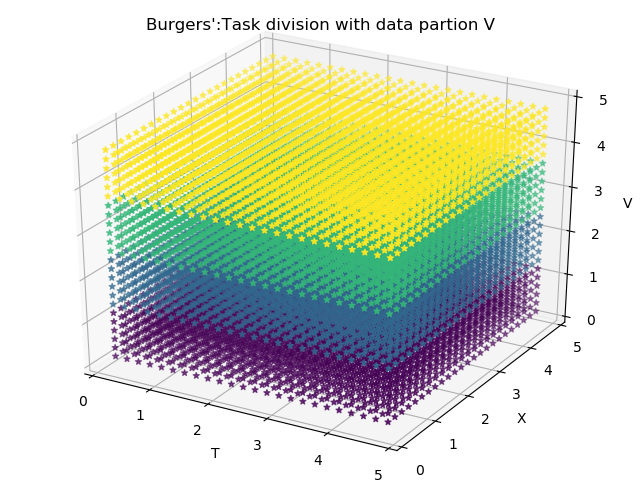}
}
\quad
\subfigure[]{
\includegraphics[scale=0.2]{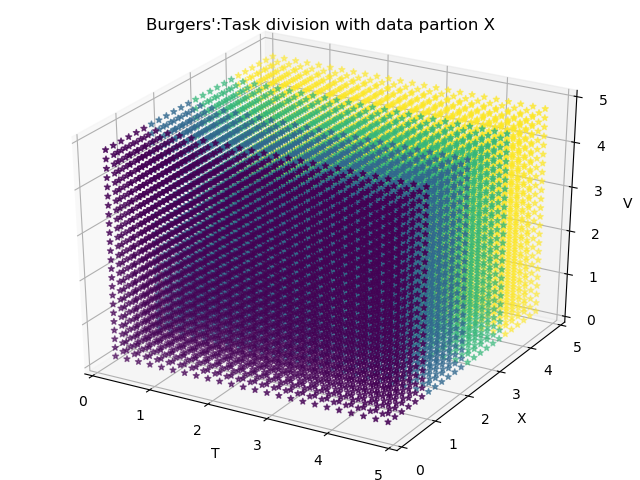}
}
\caption{The results of task allocation in Burgers' experiment. Subgraph (a) denotes the results of k-means, (b) denotes the results of data partition by $v$ and (c) denotes the results of data partition by $x$. }\label{fig_burgers_task_division_visualization}
\centering
\end{figure}

\begin{figure}[htbp]
\centering
\includegraphics[scale=0.3]{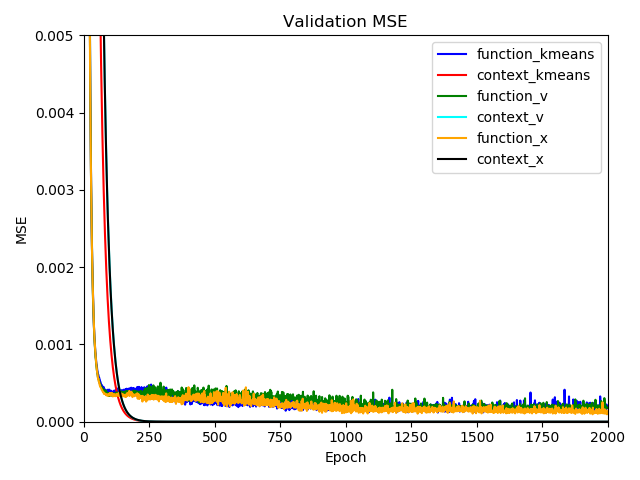}
\caption{The loss variations of function network and context network of MTL\_K-means, MTL\_v and MTL\_x in Burgers' experiment. The 'function\_kmeans' and 'context\_kmeans' denote the loss function of the function networks and the context networks from MTL\_K-means, respectively. The rest of signs are similar.}\label{fig_loss_function_burgers}
\centering
\end{figure}

Fig.\ref{fig_burgers_task_division_visualization} shows the visualization results of task allocation. Fig.\ref{fig_loss_function_burgers} depicts the variations of loss functions of three MTLs. Apparently, the loss variations of function networks are similar, as well as the context networks. Besides, the loss value of context networks is close to 0, which implies that all three MTLs learn aerodynamic characteristics strictly according to the subtasks showed in the Fig.\ref{fig_burgers_task_division_visualization}.

\begin{figure}[htbp]
\centering
\subfigure[]{
\includegraphics[scale=0.14]{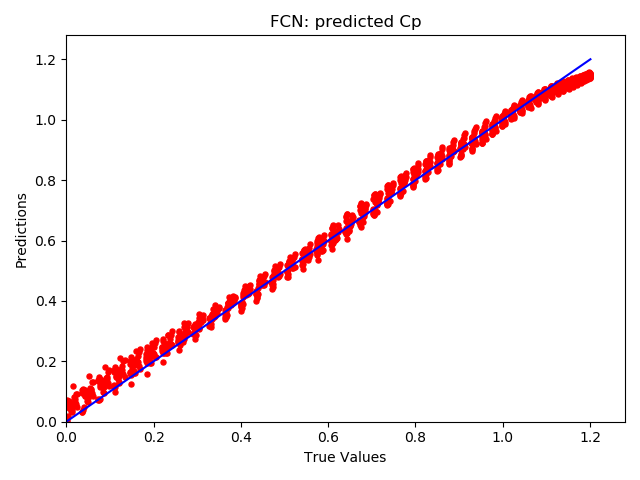}
}
\quad
\subfigure[]{
\includegraphics[scale=0.14]{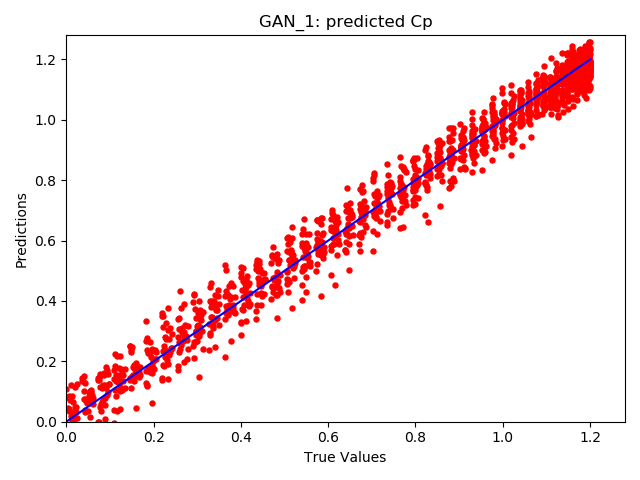}
}
\quad
\subfigure[]{
\includegraphics[scale=0.14]{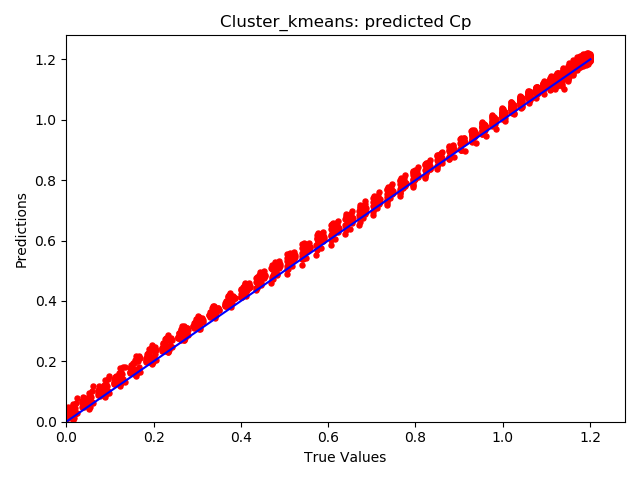}
}
\quad
\subfigure[]{
\includegraphics[scale=0.14]{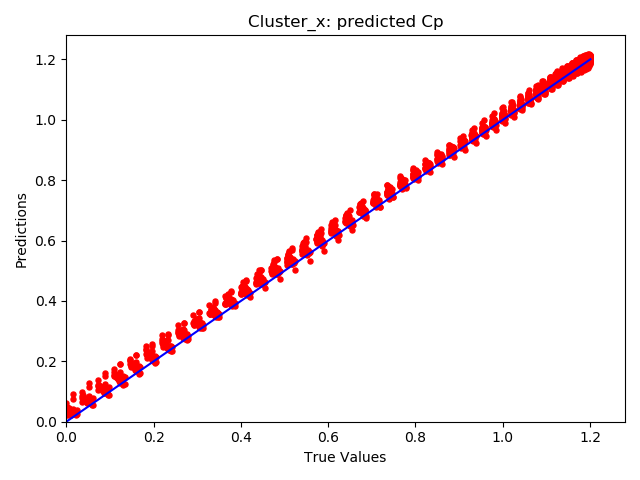}
}
\quad
\subfigure[]{
\includegraphics[scale=0.14]{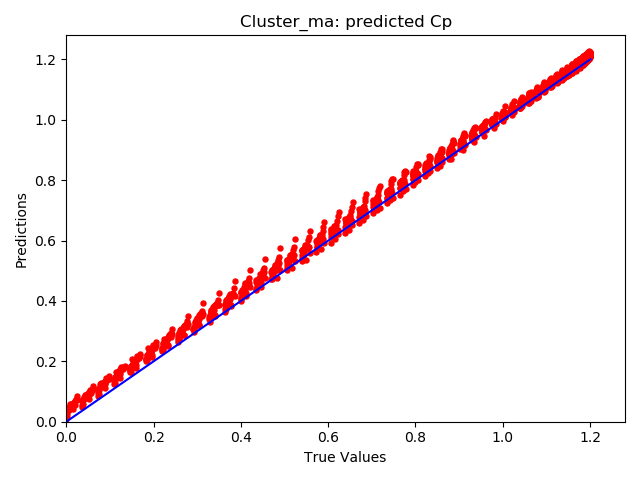}
}

\caption{The $C_{p}$ predicted by FCN\_1 (a), GAN\_1 (b), MTL\_k-means (c), MTL\_x (d) and MTL\_Ma (e).}\label{fig_ns_predicted_cp}
\centering
\end{figure}

\subsubsection{Cylindrical Laminar Experiment}

In cylindrical laminar experiment, we test the same approaches, whose test errors are shown in Tab.\ref{tab_ns_models_compare}. We can learn that all MTLs are better than others in terms of MSE and MAE. Besides, the basis of task allocation do not affects the test errors apparently.

\begin{table}[tbh]
\centering
\begin{tabular}{cccc}
\hline
Method & structure & MSE & MAE\tabularnewline
\hline
FCN\_1 & 3{*}32 & $1.31\times10^{-4}$ & $8.44\times10^{-3}$\tabularnewline
FCN\_2 & 3{*}64 & $2.45\times10^{-4}$ & $8.81\times10^{-3}$\tabularnewline
GAN\_1 & G(62,1{*}64,4)D(4,1{*}64,1) & $6.35\times10^{-4}$ & $1.91\times10^{-2}$\tabularnewline
GAN\_2 & G(62,3{*}64,4)D(4,3{*}64,1) & diverges & diverges\tabularnewline
MTL\_K-means & 4;3{*}64;1{*}5 & $8.51\times10^{-5}$ & $6.80\times10^{-3}$\tabularnewline
MTL\_x & 4;3{*}64;1{*}5 & $8.01\times10^{-5}$ & $6.28\times10^{-3}$\tabularnewline
MTL\_Ma & 4;3{*}64;1{*}5 & $7.65\times10^{-5}$ & $6.32\times10^{-3}$\tabularnewline
\hline
\end{tabular}
\caption{The test errors of cylindrical laminar experiment.}\label{tab_ns_models_compare}
\end{table}

\begin{figure}[htbp]
\centering
\subfigure[]{
\includegraphics[scale=0.14]{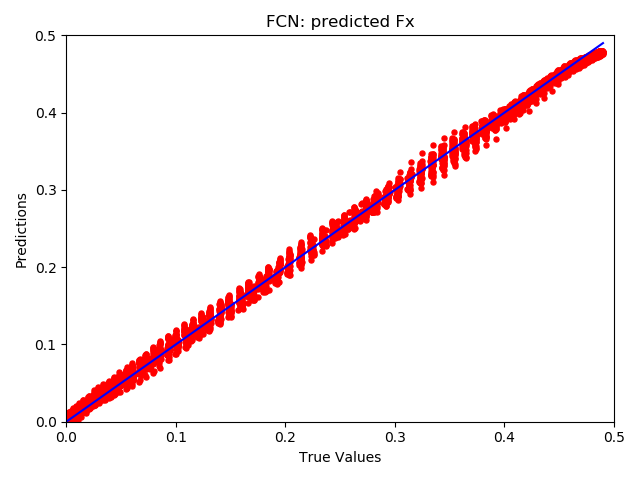}
}
\quad
\subfigure[]{
\includegraphics[scale=0.14]{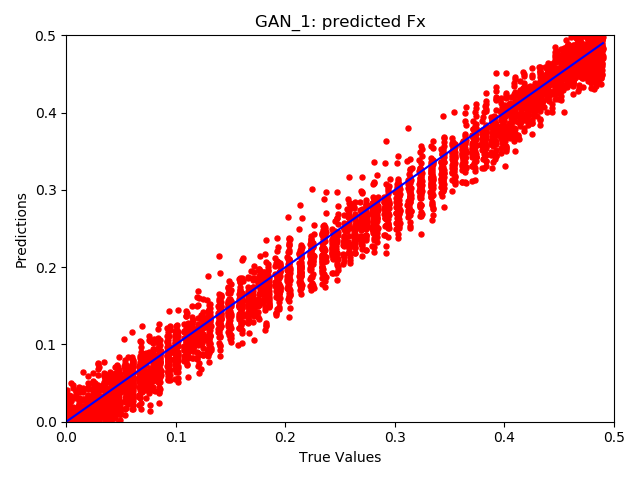}
}
\quad
\subfigure[]{
\includegraphics[scale=0.14]{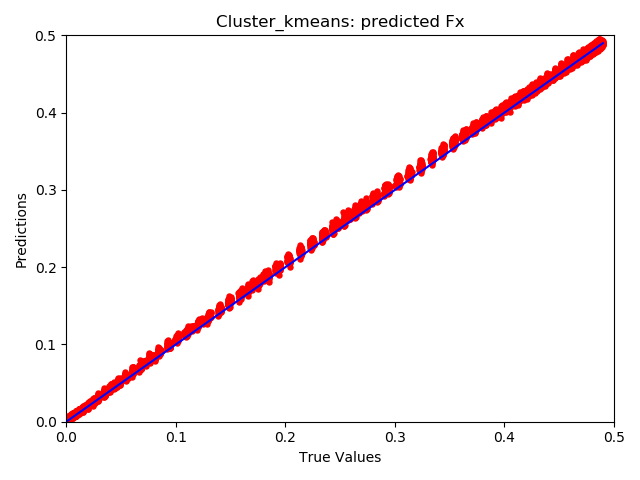}
}
\quad
\subfigure[]{
\includegraphics[scale=0.14]{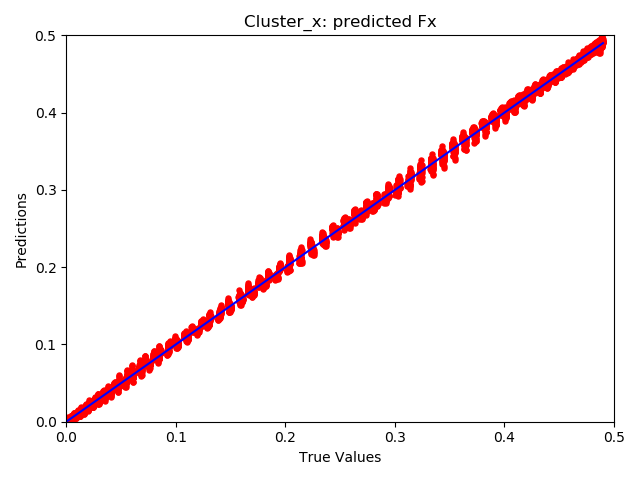}
}
\quad
\subfigure[]{
\includegraphics[scale=0.14]{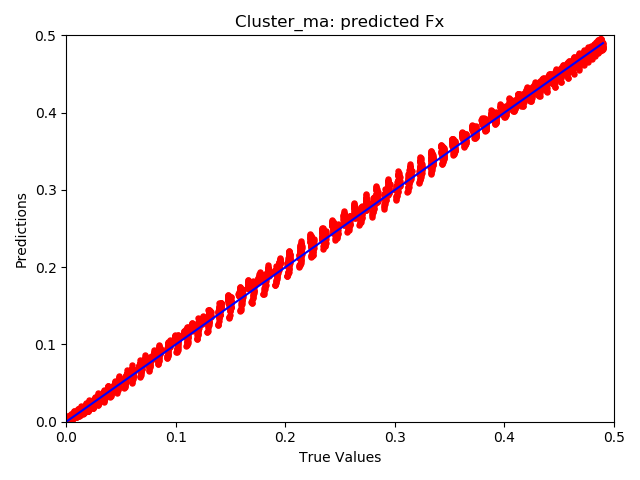}
}

\caption{The $F_{x}$ predicted by FCN\_1 (a), GAN\_1 (b), MTL\_k-means (c), MTL\_x (d) and MTL\_Ma (e).}\label{fig_ns_predicted_fx}
\centering
\end{figure}

The visualization results of $C_{p}$ and $F_{x}$ predicted by the above approaches are shown in Fig.\ref{fig_ns_predicted_cp} and Fig.\ref{fig_ns_predicted_fx}. Like Burgers' experiment, the FCN still can not predicted $C_{p}$ and $F_{x}$ precisely where their values are relatively large. In addition, the cGAN could not achieve accurate prediction of $C_{p}$ and $F_{x}$ in the entire flow field. On the contrary, all the three MTLs can achieve accurate prediction of both $C_{p}$ and $F_{x}$ in the entire flow field.

\begin{figure}[htbp]
\centering
\subfigure[]{
\includegraphics[scale=0.12]{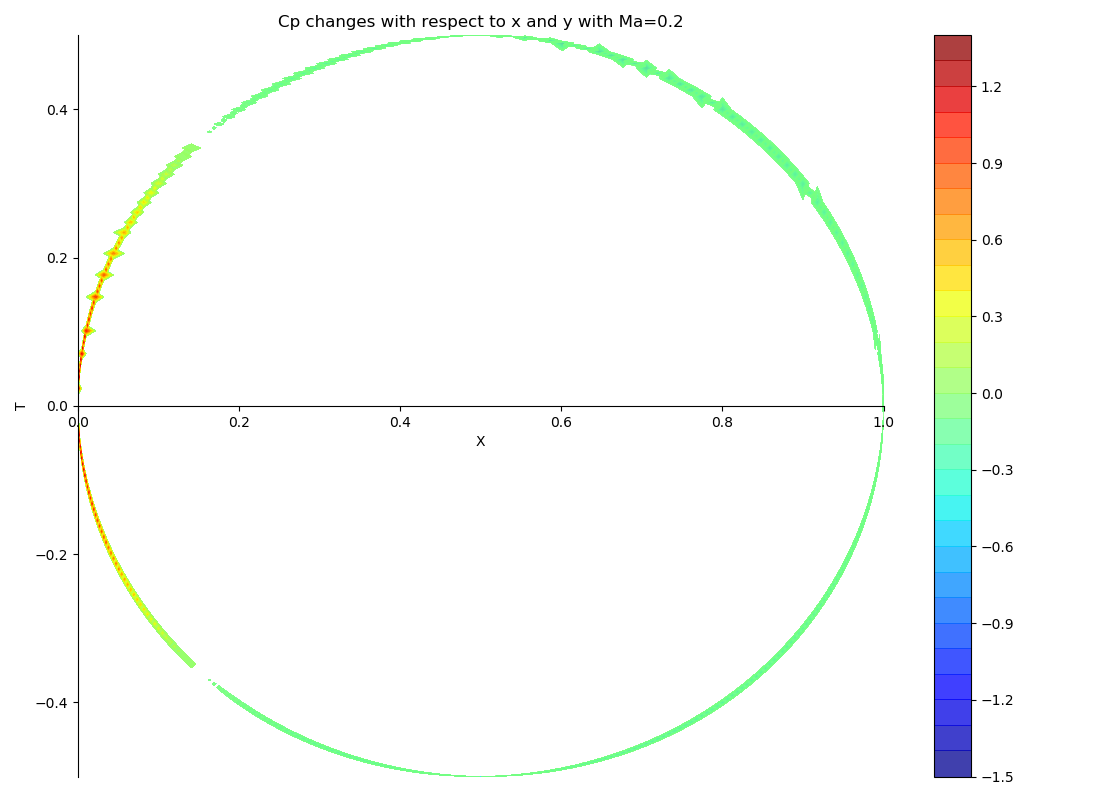}
}
\quad
\subfigure[]{
\includegraphics[scale=0.12]{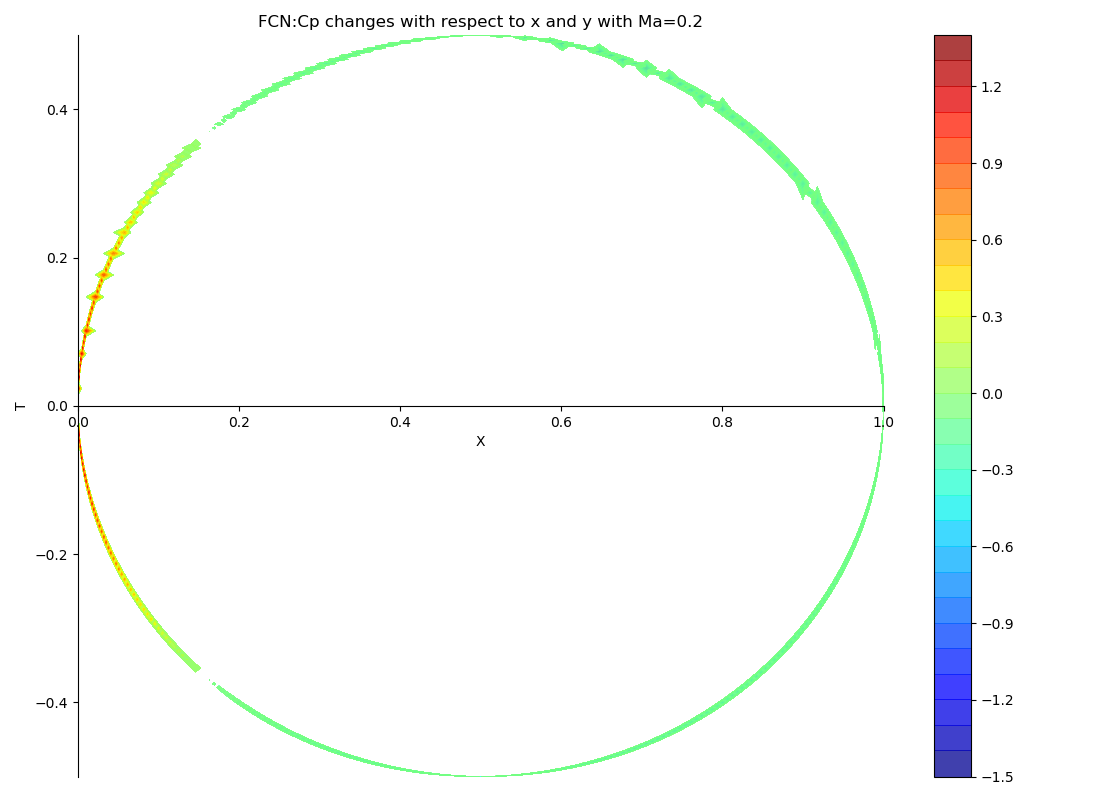}
}
\quad
\subfigure[]{
\includegraphics[scale=0.12]{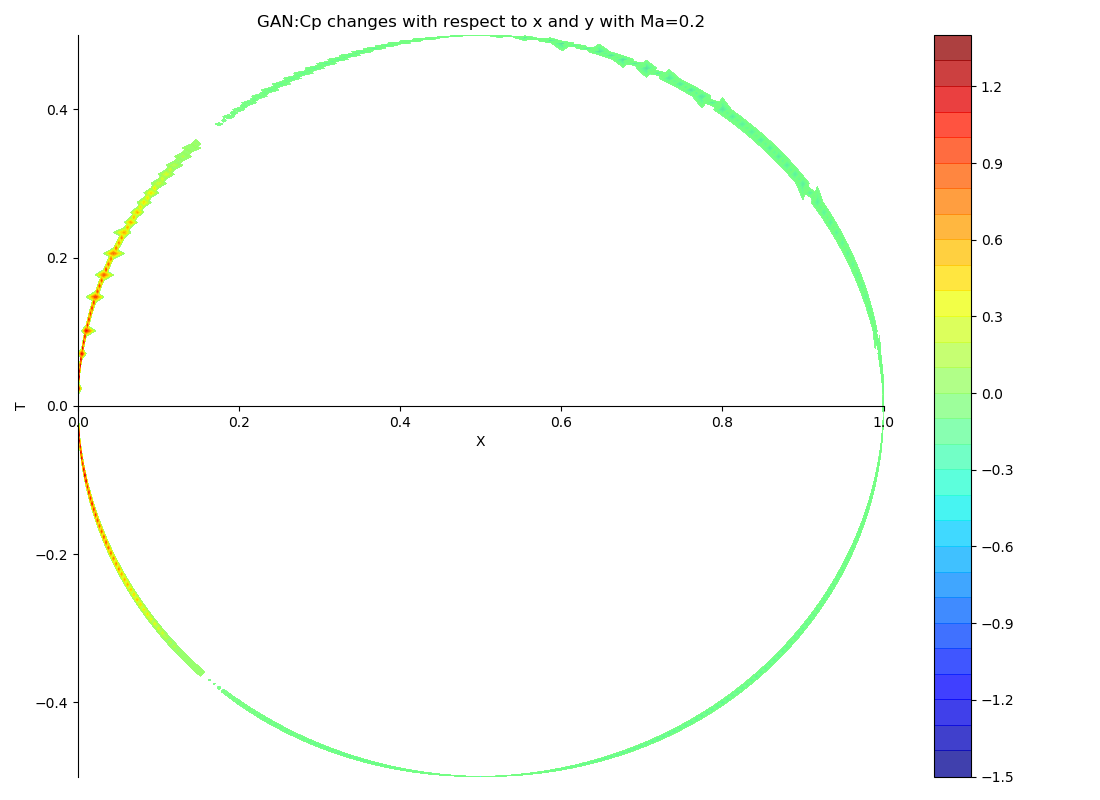}
}
\quad
\subfigure[]{
\includegraphics[scale=0.12]{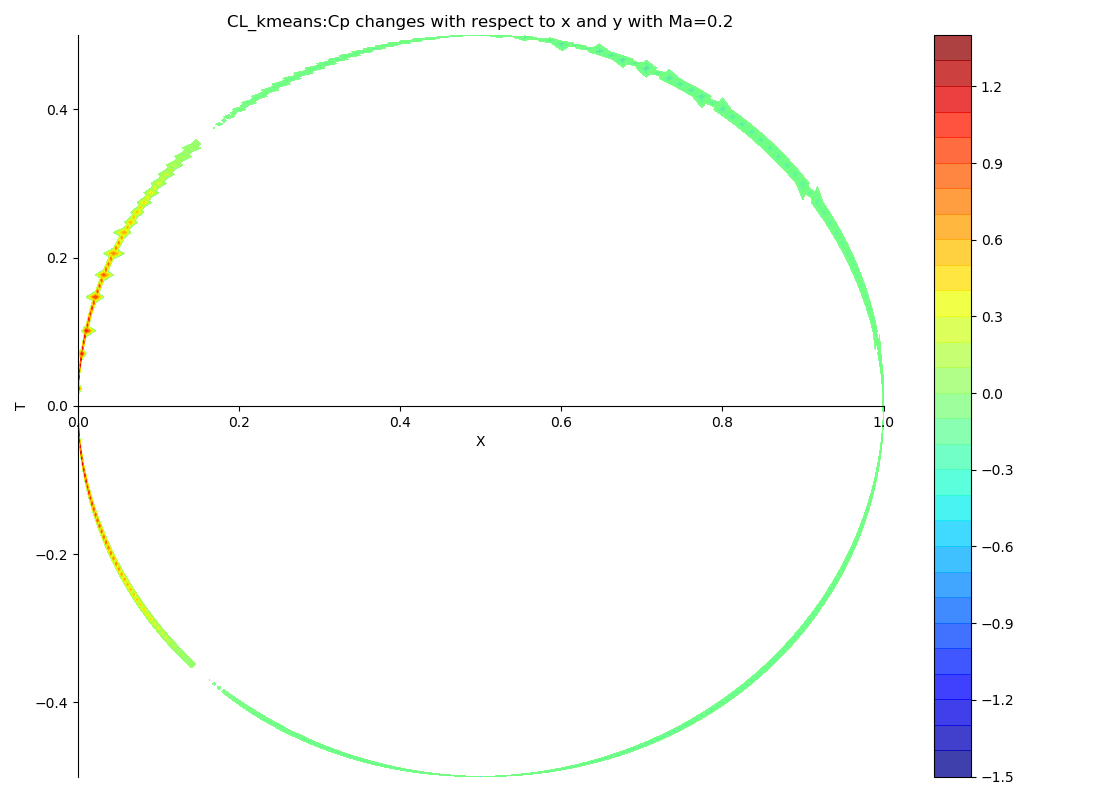}
}
\quad
\subfigure[]{
\includegraphics[scale=0.12]{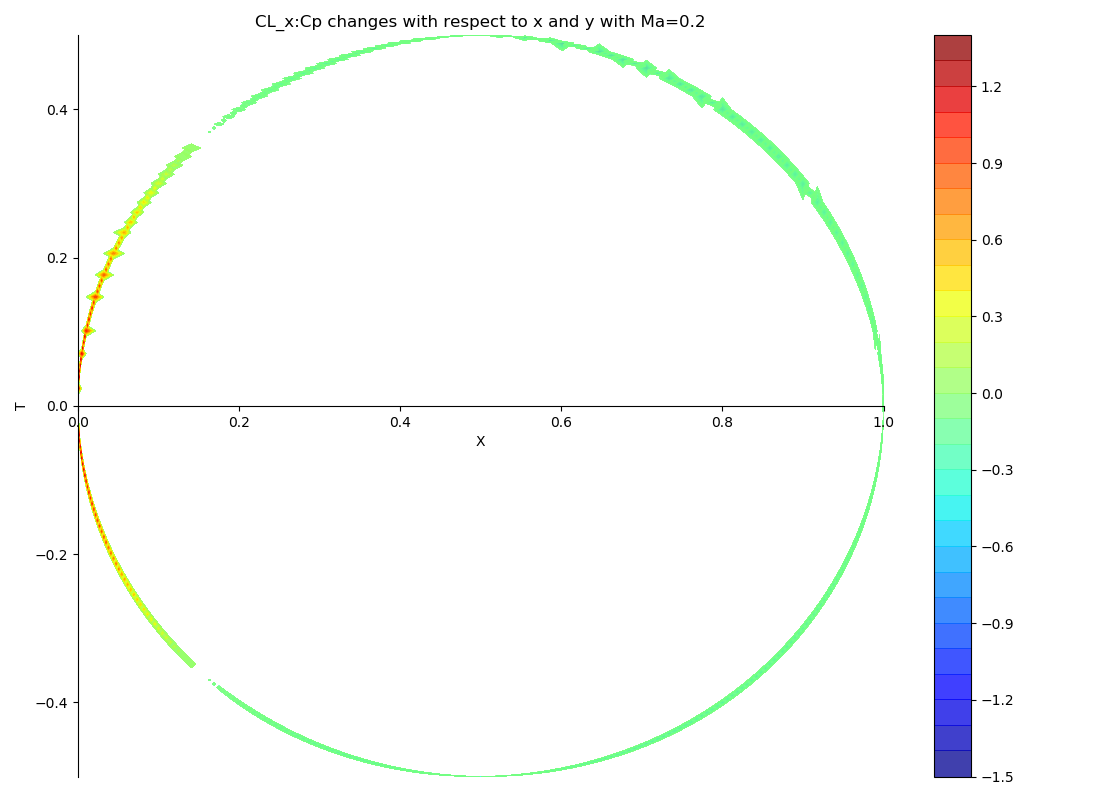}
}
\quad
\subfigure[]{
\includegraphics[scale=0.12]{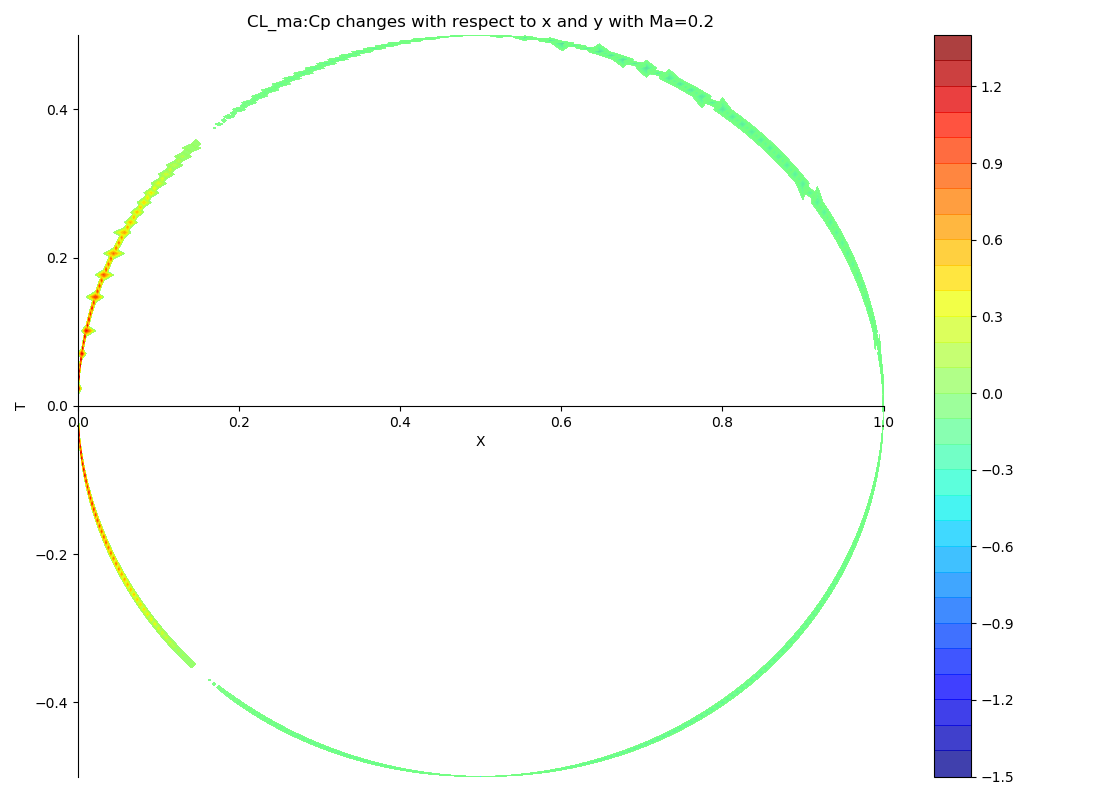}
}
\caption{The $C_{p}$ predicted by CFD (a), FCN\_1 (b), cGAN\_1 (c), MTL\_k-means (d), MTL\_v (e) and MTL\_x (f).}\label{fig_ns_visualization}
\centering
\end{figure}

\begin{figure}[htbp]
\centering
\subfigure[]{
\includegraphics[scale=0.3]{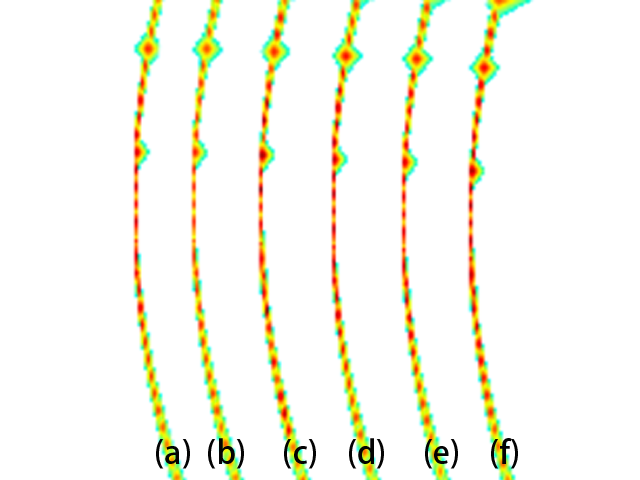}
}
\quad
\subfigure[]{
\includegraphics[scale=0.3]{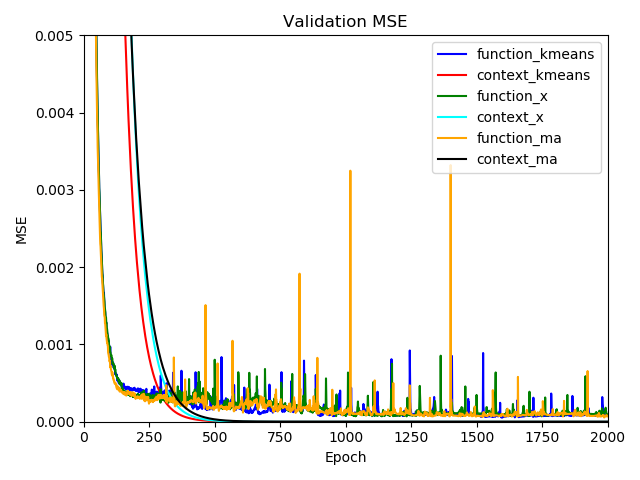}
}
\caption{Subgraph (a) denotes the enlargement of experimental visualization results in Fig.\ref{fig_ns_visualization}. Subgraph (b) denotes the loss variations of function network and context network of MTL\_K-means, MTL\_x and MTL\_Ma in cylindrical laminar experiment. }\label{fig_miss}
\centering
\end{figure}

The visualization results of $Cp$ predicted by the above approaches are shown as in Fig.\ref{fig_ns_visualization}. Intuitively, all these models can predict the $Cp$ around the surface of a cylindrical. However differences can be seen clearly after zooming in those subgraphs. We enlarge the left half of the cylinder (i.e. the flow field with a large $C_{p}$) in Fig.\ref{fig_miss}(a). We can see that the red part comes from subgraph (b) in Fig.\ref{fig_ns_visualization} is lighter than others, which is consistent with the conclusion derived from Fig.\ref{fig_ns_predicted_cp}

\begin{figure}[htbp]
\centering
\subfigure[]{
\includegraphics[scale=0.2]{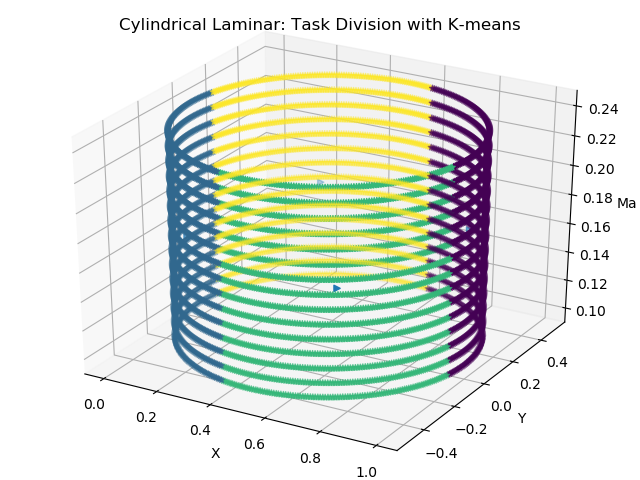}
}
\quad
\subfigure[]{
\includegraphics[scale=0.2]{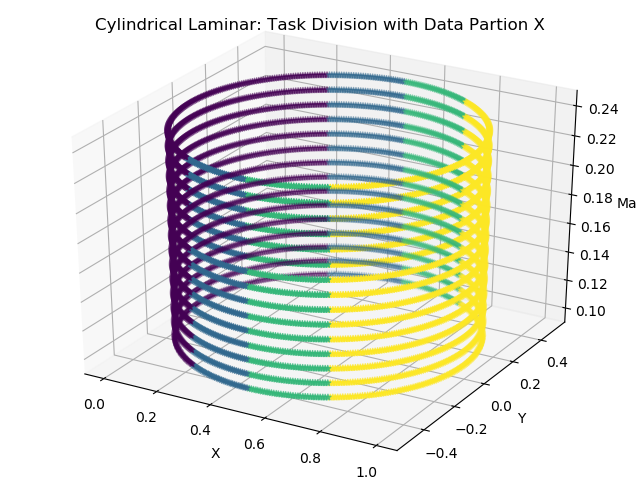}
}
\quad
\subfigure[]{
\includegraphics[scale=0.2]{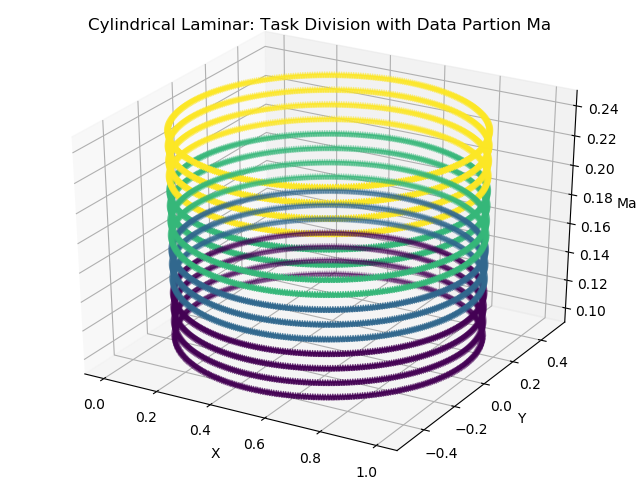}
}
\caption{The results of task allocation in cylindrical laminar experiment. Subgraph (a) denotes the results of k-means, (b) denotes the results of data partition by $x$ and (c) denotes the results of data partition by $Ma$.}\label{fig_ns_task_division_visualization}
\centering
\end{figure}

Fig.\ref{fig_ns_task_division_visualization} depicts the visualization results of task allocation. Fig.\ref{fig_miss}(c) shows the variations of loss functions of three MTLs. Also, the loss variations of context networks are similar and close to 0. However, the fluctuations of function networks are different. As we can see, the fluctuations of function loss of the MTL which divides subtasks by $Ma$ is more drastic than others.

\subsection{Analysis}

\begin{table}[htb]
\centering
\begin{tabular}{cccccccccccc}
\hline
Sample Id &$v$  & real $u$ & predicted $u$ &$f_{1}$ & $c_{1}$      &$f_{2}$ &$c_{2}$       &$f_{3}$ &$c_{3}$       &$f_{4}$ &$c_{4}$\tabularnewline
\hline
1         & 1.0 & 0.69     & 0.68          &0.09    &\textbf{0.98} &0.14    &0.48          &0.00    &0.00          &0.00    &0.00\tabularnewline
2         & 1.8 & 2.79     & 2.08          &0.62    &0.00          &0.96    &\textbf{0.48} &0.89    &0.01          &0.81    &0.00\tabularnewline
3         & 3.0 & 3.01     & 2.11          &0.06    &0.00          &0.88    &0.00          &0.97    &\textbf{0.49} &0.80    &0.00\tabularnewline
4         & 3.8 & 4.04     & 3.77          &0.00    &0.00          &0.87    &0.00          &0.94    &0.49          &0.40    &\textbf{0.99}\tabularnewline
\hline
\end{tabular}
\caption{The output of each cluster of MTL\_v in Burgers' experiment. In the column related to "real $u$" and "predicted $u$", the data are not normalized, however the data in the rest of columns are normalized. Bold numbers means that the corresponding cluster is activated.}\label{tab_Burgers_mid_result}
\end{table}

We take Burgers' experiment as an example to analyze the data quality-adaptivity of MTLs. Tab.\ref{tab_Burgers_mid_result} shows four samples in the test set. We can see that sample 1 activates the first cluster in the MTL\_v, sample 2 activates the second cluster, and so on. It is obvious that samples with different $v$ are processed by different clusters in the MTL\_v. Therefore, the MLT\_v distributes all learning pressure on every cluster, so that a cluster could focus on a small learning task and the learning process among clusters does not affect each other. This mechanism is not affected by the quality of the dataset, i.e. it is an dataset quality-adaptive learning scheme, which perfectly explains why the MTLs are more accurate than FCNs and GANs in entire flow field.

\section{Conclusions}

In this paper, the MTL in aerodynamic data modeling is proposed. We first pointed out the impact of the quality of datasets on the accuracy of aerodynamic data models. Then, we propose the MTL, a dataset quality-adaptive learning scheme, which consists of tasks allocation and aerodynamic characteristics learning. Finally, the dataset quality-adaptivity of the MTL is verified through Burgers' and cylinder laminar experiments. Compared with existing models, MTL possess the ability of adapting to poor quality datasets.

The detailed conclusions are as follows:

a) the MTL is a dataset quality-adaptive learning scheme;

b) the MTL is more accuracy than FCNs and GANs;

c) the advantages described in a) and b) come from the partial activation mechanism of the MTL.

\bibliographystyle{unsrt}
\bibliography{main}

\end{document}